\documentclass[lettersize,journal]{IEEEtran}
\usepackage{amsmath,amsfonts}
\usepackage{algorithmic}
\usepackage{array}
\usepackage{subcaption}
\usepackage{textcomp}
\usepackage{stfloats}
\usepackage{url}
\usepackage{verbatim}
\usepackage{graphicx}
\usepackage{cite}
\usepackage{caption}       
\usepackage{overpic}       
\usepackage{xcolor}        
\usepackage{multicol}      
\def\eps{{\epsilon}}
\usepackage{amssymb}
\usepackage{subcaption}
\usepackage{booktabs}   
\usepackage{multirow}   

\definecolor{iccvblue}{rgb}{0.21,0.49,0.74}
\usepackage[pagebackref,breaklinks,colorlinks,allcolors=iccvblue]{hyperref}
\usepackage{xcolor,colortbl}
\usepackage{pifont}
\newcommand{\xmark}{\ding{55}}%
\newcommand{\maxf}[1]{{\cellcolor[gray]{0.8}} #1}
\definecolor{my_green}{rgb}{0.302, 0.686, 0.290}
\definecolor{my_red}{rgb}{0.894, 0.102, 0.110}

\usepackage[ruled,vlined]{algorithm2e}
\usepackage[dvipsnames]{xcolor}

\title{MUSE: Manipulating Unified Framework for Synthesizing Emotions in Images via Test-Time Optimization}

\author{
~~Yingjie Xia$^{1}$~~~~~~~~~Xi Wang$^{2}$~~~~~~~~~~Jinglei Shi$^{1,3}$~~~~~~~~Vicky Kalogeiton$^{2}$~~~~~~~~
Jian Yang$^{1}$\\
$^1$ VCIP \& TMCC \& DISSec, College of Computer Science, Nankai University\\
$^2$ LIX, Ecole Polytechnique, IP Paris~~~ 
$^3$ Key Lab of SCCI, Dalian University of Technology
}

\begin{document}

\maketitle

\begin{abstract}
Images evoke emotions that profoundly influence perception, often prioritized over content. Current Image Emotional Synthesis (IES) approaches artificially separate generation and editing tasks, creating inefficiencies and limiting applications where these tasks naturally intertwine, such as therapeutic interventions or storytelling. 
In this work, we introduce MUSE, the first unified framework capable of both emotional generation and editing. 
By adopting a strategy conceptually aligned with Test-Time Scaling (TTS) that widely used in both LLM and diffusion model communities, it avoids the requirement for additional updating diffusion model and specialized emotional synthesis datasets.
More specifically, MUSE addresses three key questions in emotional synthesis: 
(1) \textbf{\textit{HOW}} to stably guide synthesis by leveraging an off-the-shelf emotion classifier with gradient-based optimization of emotional tokens; 
(2) \textbf{\textit{WHEN}} to introduce emotional guidance by identifying the optimal timing using semantic similarity as a supervisory signal; 
and (3) \textbf{\textit{WHICH}} emotion to guide synthesis through a multi-emotion loss that reduces interference from inherent and similar emotions. 
Experimental results show that MUSE performs favorably against all methods for both generation and editing, improving emotional accuracy and semantic diversity while maintaining an optimal balance between desired content, adherence to text prompts, and realistic emotional expression. 
It establishes a new paradigm for emotion synthesis.
\end{abstract}

\section{Introduction}
\textit{``Conscience is the voice of the soul, the passions are the voice of the body."\hfill -Jean-Jacques Rousseau (FR.)} \\
\IEEEPARstart{M}{edia}, particularly images, have profoundly transformed the way people perceive the world in dissemination. 
People sometimes prioritize the emotion conveyed in images over content, a phenomenon so impactful that gives rise to the ``post-truth" concept in political science.
This emotional impact has led to fast progress in Image Emotional Synthesis (IES) across various
fields~\cite{van2012design,mann2005emotional}.

Current IES work typically splits into two streams: (1) \emph{generation}, which creates novel imagery with a desired emotional profile, and (2) \emph{editing}, which retouches existing images to alter their affect. Maintaining \textit{separate} pipelines for these tasks inflates training cost, complicates deployment, and most critically, introduces inconsistency between the generation and refinement. A unified framework would collapse that divide, enabling continuous cycles of generation and adjustment that are vital for therapeutic art tools, adaptive storytelling engines, and emotion-aware communication platforms where creation and modification blur in practice.

Early IES methods conveyed emotions by editing low-level features, such as linking colors to affective words~\cite{wang2013affective} or applying color constraints via emotion classifiers~\cite{liu2018emotional}. Later approaches explored high-level styles, using semantic spaces~\cite{sun2023msnet} or text-driven color guidance~\cite{weng2023affective}.
Recent IES approaches leverage diffusion~\cite{rombach2022high,brooks2023instructpix2pix} 
following its success in various content generation.
EmoGen~\cite{yang2024emogen} pioneered emotional diffusion generation by mapping emotion and semantic spaces, translating abstract emotions into concrete attributes for the generation task.
EmotiCrafter~\cite{he2025emoticrafter} extends emotional generation to the continuous valence-arousal space by embedding emotions into text features. 
Emotional editing has also been advanced: 
EmoEdit~\cite{yang2025emoedit} first employs GPT4V to extract emotion factor trees that identify key emotional attributes.
Recently, EmoAgent~\cite{mao2025emoagent} employs large language models (LLMs) to generate editing instructions and calls specific editing tools (IP2P~\cite{radford2021learning}) 
for affective image manipulation.

Despite these advances, SOTA IES methods continue to face several key limitations: %
(1) The color- and style-based approaches~\cite{yang2008automatic,zhu2023emotional,sun2023msnet,weng2023affective} often fail to accurately evoke specific emotions due to the strong inherent image emotion.
(2) While the diffusion-based methods~\cite{lin2024make} often rely on specialized datasets, and may significantly change the content or even produce a totally different image. 
(3) Methods that implicitly~\cite{yang2024emogen} or explicitly~\cite{yang2025emoedit} associate emotions with certain attributes will make some elements repeatedly appear in the synthesized images, reducing their semantic diversity. 
Finally, (4) existing approaches address either emotional generation~\cite{yang2024emogen,lin2024make,he2025emoticrafter} or editing~\cite{yang2025emoedit,zhu2023emotional}, creating inconsistent emotional representations and fragmenting user experience.

To address these issues, we propose MUSE, a unified framework for both emotional generation and editing (see Fig.\ref{fig:main}). 
Our idea is similar to the test-time scaling philosophy\cite{ma2025inference}, we search for a better noise via reverse optimization of tokens during inference to achieve more precise and harmonious emotion control.
It focuses on 
\emph{how, when}, and \emph{which} emotions should guide the synthesis:
(1). \textbf{How:} 
We leverage prior knowledge from an off-the-shelf emotion classifier to guide the synthesis process. 
Gradients are back-propagated to progressively optimize predefined emotional tokens at the input. 
As a result, a longer denoising process is employed during inference, but leading to improved synthesis quality.
(2). \textbf{When:} 
While scaling during the inference stage is beneficial, introducing emotional guidance too early or too late may disrupt image content or fail to convey the target emotion. 
To address this, we use semantic similarity as a supervisory signal to determine the optimal timing for applying guidance, striking a balance between effective control and efficient generation.
(3). \textbf{Which:} We propose a multi-emotion cross-entropy loss to reduce interference from both the inherent and similar emotions, improving the emotional synthesis accuracy and alleviating conflicts between the target and inherent emotions.
 
In summary, MUSE optimizes emotional tokens by minimizing a cross-entropy emotional loss, then introduces these tokens at the optimal timing to guide image emotional synthesis.
Extensive experiments and analysis show that MUSE strikes a notable balance among desired content, textual adherence and emotional realism. 
It achieves superior emotion accuracy and human preference across both tasks against those SOTA emotion generation methods such as UG~\cite{bansal2023universal}, SD3~\cite{esser2024scaling}, FLUX~\cite{flux2024},
EmoGen~\cite{yang2024emogen}, 
and emotion editing ones like SDEdit (SDE)~\cite{mengsdedit}, AIF~\cite{weng2023affective}, IDE~\cite{zou2024towards}, PnP~\cite{tumanyan2023plug}, Forgedit~\cite{zhang2023forgedit}, and EmoEdit~\cite{yang2025emoedit},
\emph{without} relying on specialized datasets or additional updates of the diffusion model. 
The contributions of our work can be summarized as follows: 
\begin{itemize}
    \item We propose the first unified framework for both emotional image generation and editing, which leverages a test-time optimization strategy to achieve more accurate and natural emotion manipulation without creating specialized dataset and model finetuning.
    \item 
    We investigate the three key aspects of how, when, and which emotion to guide the synthesis by continuously optimizing emotional tokens. CLIP-based semantic similarity is employed to ensure alignment between generated images and text prompts, while a multi-emotion loss function guides the accurate expression of the target emotions.
    These factors together ensure a stable synthesis.
    \item 
    We carried out comprehensive experiments on both emotional editing and generation tasks to demonstrate that MUSE outperforms recent State-Of-The-Art (SOTA) methods such as EmoGen~\cite{yang2024emogen} and EmoEdit~\cite{yang2025emoedit} in terms of emotional accuracy (Emo-A/B/C), 
    and visual realism (FID~\cite{heusel2017gans}), while also ensuring high semantic similarity (CLIP) and human preference (HPSv2~\cite{wu2023human}).
    In addition, we reported superior performance in corresponding human evaluation studies against compared methods.
    
\end{itemize}

\section{Related work}
\subsection{Diffusion Models for Image Synthesis}
Diffusion models for image synthesis gradually add random noise to data and learn a model to reverse this process to recover data from the original distribution~\cite{sohl2015deep,ho2020denoising}. 
Methods either operate in pixel space, e.g.
Imagen~\cite{saharia2022photorealistic} or in latent space for improved efficiency at higher resolutions such as DALLE-2~\cite{ramesh2022hierarchical}, Stable Diffusion (SD)~\cite{rombach2022high}.
To condition the generation, most works apply guidance~\cite{ho2021classifier,dhariwal2021diffusion} during denoising.
The authors in~\cite{ho2021classifier} introduce classifier-free guidance to guide the generation directly without an explicit pre-trained classifier.
To further enhance controllability, recent works explore various plug-and-play modules. 
For example, IP-Adapter~\cite{ye2023ip} injects features from reference images via cross-attention operation
Moreover, structure-aware controllers such as ControlNet~\cite{zhang2023adding} and T2I-Adapter~\cite{mou2024t2i} incorporate spatial conditions like edges or depth maps to guide image layout.
More recently, ~\cite{ma2025inference} proposed a test-time scaling framework that improves generation quality by searching for better noise inputs during inference, complementing the traditional denoising step scaling.

\subsection{Visual Emotion Analysis}
Visual Emotion Analysis (VEA) traditionally focused on designing handcrafted features for analyzing emotions in images~\cite{zhao2014affective}.
Modern works~\cite{yang2018visual,zhang2019exploring,xu2022mdan} rely on DL for emotional recognition.
Later works~\cite{deng2022simple,pan2024multi} leveraged LLMs. 
They either proposed prompt-based fine-tuning to improve classification~\cite{deng2022simple}, or proposed language-supervised fusion of language and emotion features to boost the emotion capability of visual models~\cite{pan2024multi}.
Emotion analysis has also been explored in videos~\cite{zhang2024mart,zhang2018recognition} and online chatting~\cite{nie2023long,lu2024hypergraph}.
~\cite{zhang2024mart} proposed a Masked AutoEncoder-based video affective analysis method, employing lexicon verification and emotion-oriented masking for robust temporal emotion representation learning.
For conversational emotion recognition, ~\cite{ma2023transformer} introduced a transformer-based model with self-distillation to enhance multimodal interactions and representations. In addition to recognition methods, ~\cite{zhao2024to} introduced two psychological metrics (ECC and EMC) to quantify the severity of emotional misclassification, emphasizing emotion-aware evaluation.

\begin{figure*}[!th]
  \centering
  \includegraphics[width=0.99\textwidth]{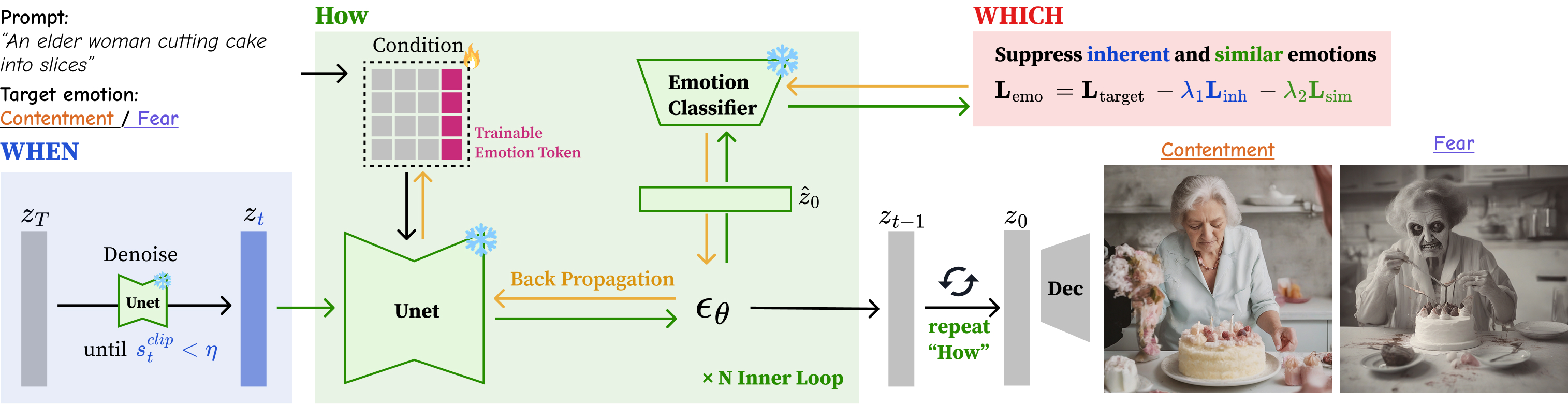}
  \caption{
 Our unified framework, MUSE, addresses the \textit{\textcolor{Green}{how}}, \textit{\textcolor{blue}{when}},  and \textit{\textcolor{red}{which}} questions in emotional image synthesis. It starts from a noise latent \( z_T \) that is either sampled from a Gaussian distribution (for generation) or obtained via inversion from an existing image (for editing). A CLIP model is employed to compute semantic similarity \( s^{clip}_t \) between the synthesized image and the text prompt for determining \textit{\textcolor{blue}{when}} to introduce the emotional guidance. To decide \textit{\textcolor{Green}{how}} to manipulate the emotion, we introduce a learnable emotion token in addition to the textual prompt tokens, during the inference stage. These tokens are optimized in an inner loop using an off-the-shelf emotional classifier that operates on the predicted denoised latent \( \hat{z}_0 \).
  Furthermore, we propose an emotional loss $\mathcal{L}_{\text{emo}}$ to ensure \textit{\textcolor{red}{which}} emotions to enhance, which comprises the terms $\mathcal{L}_{\text{target}}$ for synthesizing the target emotion, and $\mathcal{L}_{\text{inh}}$, $\mathcal{L}_{\text{sim}}$ for suppressing inherent and similar emotions respectively. 
  }
  \label{fig:main}
\end{figure*}

\subsection{Image Emotional Synthesis}
Image Emotional Synthesis (IES) generates media aligned with emotions through image generation and editing.
Editing methods usually rely on color~\cite{ zhu2023emotional}, style~\cite{weng2023affective} and diffusion~\cite{lin2024make, yang2025emoedit}.
The first attempt~\cite{yang2008automatic} associates color moods with images via histograms, while \cite{zhu2023emotional} disentangles neutral high-level concepts from emotional low-level features via adversarial training. 
Style-based methods alter the overall image style, as in~\cite{fu2022language}, which proposes a language-driven method transferring emotions based on style images and text instructions.
Affective Image Filter (AIF)\cite{weng2023affective} employs a multi-modal transformer to translate abstract emotional cues from text into concrete visual elements such as color and texture.
Diffusion models edit image emotions by altering the semantics of their content. \cite{yang2025emoedit} introduces emotion factor trees to identify attributes requiring modification by Vision-Language Models (VLMs) for emotion adjustment. 
Similarly, \cite{yang2024emogen} pioneers emotional generation by linking emotions to specific VLM-focused attributes, but both works suggest that emotions are tied to fixed elements, thereby limiting semantic diversity.
The work of Make Me Happier (MMH)\cite{lin2024make} proposes a diffusion-based editing framework that preserves image semantics while modifying emotions, though it relies on simple emotion encodings that may limit expressive richness.
EmotiCrafter~\cite{he2025emoticrafter} embeds Valence-Arousal into text features but struggles with arousal control and suffers from human-centric bias.
EmoAgent~\cite{mao2025emoagent} introduces a multi-agent framework that plans, edits, and critiques emotional attributes in a structured manner, but increases system complexity and relies on predefined emotional attributes.
Despite these advances, a flexible framework handling both tasks with high emotional accuracy and quality remains absent, a gap our work addresses.
\section{Preliminary Knowledge}
Diffusion for image synthesis involves reverting random noise to an image that follows the data distribution. 
It involves a forward addition of noise and a backward denoising process.

In the forward process, an image \(I_0\) is initially processed through a VAE-like encoder to obtain a latent embedding \(z_{0} = \mathcal{E}(I_{0})\). This embedding is then progressively destroyed by adding random noise over \(T\) steps:
\begin{equation}
z_t = \sqrt{\alpha_t} z_0 + \sqrt{1 - \alpha_t} \epsilon, \quad \epsilon \sim \mathcal{N}(0, I) \quad,
\label{eq1:forward_pass}
\end{equation}
where \(t \in [1, T]\) represents the time step, and \(\{\alpha_l\}_{l=1}^T\) is a scheduler that gradually decreases, controlling the extent of noise mixed with the data.
In the backwards process, a network \(\epsilon_{\theta}\) is trained to reverse this diffusion process by predicting the added noise at each time step \(t\):
\begin{equation}
\epsilon_\theta\left(z_t, t, c\right) \approx \frac{z_t - \sqrt{\alpha_t} z_0}{\sqrt{1 - \alpha_t}} \quad,
\label{eq2:noise_prediction}
\end{equation}
where \(c\) is the text embedding derived from the text prompt \(P\) via the text encoder: \(c = \mathcal{T}(P)\). Then the denoised embedding at time step \(t-1\) can then be computed from the embedding at time step \(t\) as follows:
\begin{equation}
z_{t-1} = \frac{z_t - \left(\sqrt{1 - \alpha_t}\right) \epsilon_\theta\left(z_t, t, c\right)}{\sqrt{\alpha_t}} \quad.
\label{eq3:denoise_process}
\end{equation}
The network parameters \(\theta\) are optimized by minimizing the Mean Squared Error (MSE) between the predicted noise and the ground truth~\cite{ho2020denoising}:
\begin{equation}
\theta^{'} = \arg\min_\theta \mathbb{E}_{z_0, t, \epsilon \sim \mathcal{N}(0,1)}\left[\left\|\epsilon_\theta\left(z_t, t, c\right) - \epsilon\right\|^2\right] \quad.
\end{equation}
Once trained, the initial embedding distribution can be retrieved from a noise sample \(z_{T} \sim q(z_{T})\) in an iterative manner by Eq.~\ref{eq3:denoise_process}.
To apply condition information $c$ into the generation, \cite{dhariwal2021diffusion} propose \emph{Classifier Guidance (CG)} that uses a pretrained classifier $p(c|x_t)$ to disturb the inference noise $\eps_{\theta}$, with a scalar $\omega {>} 0$ controlling the guidance level of the condition $c$:
\begin{equation}
    \hat{\eps}_\theta(x_t, c) = \eps_\theta(x_t, c) + (\omega + 1) \nabla_{x_t}\log p(c|x_t) \quad.
    \label{eq: classifier guidance}
\end{equation}

To support more complex conditions, \cite{ho2021classifier} proposes to replace the classifier with an implicit classifier: by dropping the condition during training, they employ a single network for both $\nabla_{x_t}\log p(x_t, c)$ and $\nabla_{x_t}\log p(x_t)$. This gives the \emph{Classifier-Free Guidance (CFG)}, also controlled by $\omega$:
\begin{equation}
    \hat{\eps}_\theta(x_t, c) = \eps_\theta(x_t, c) + \omega \left(\eps_\theta(x_t, c) - \eps_\theta(x_t, \emptyset)\right) \quad .
    \label{eq: cfg}
\end{equation}

\section{Methodology}
We propose MUSE, a novel method for emotion-oriented image synthesis. 
As the overview shown in Fig.~\ref{fig:main}, it determines \textbf{\textit{How}}, \textbf{\textit{When}}, and \textbf{\textit{Which}} emotions to incorporate in image synthesis. 
More specifically, it takes a text prompt or source image, along with an optimizable emotional tokens as inputs for either generation task or editing task. And a target emotion is set to be used in an off-the-shelf emotion classifier (Sec.~\ref{sec:how}).
The optimization for emotional tokens begins when the semantic similarity of the synthesized image surpasses a preset threshold (Sec.~\ref{sec:when}), 
guided by a multi-emotion cross-entropy loss to ensure accurate emotion synthesis (Sec.~\ref{sec:which}).

\subsection{Image Emotional Synthesis Framework}
\label{sec:how}
\textbf{\textit{How}} to synthesize emotions during the denoising process of a diffusion model is challenging, especially with classifier-free guidance controlling textual adherence simultaneously. 
We propose using an emotion classifier as guidance to leverage its learned emotional priors.
Previous methods, such as classifier guidance~\cite{dhariwal2021diffusion} or universal guidance~\cite{bansal2023universal}, back-propagate directly to the predicted noise $\epsilon_{\theta}\left(z_{t}, t, c\right)$ as in Eq.~\ref{eq: cfg}. 
Yet, this often leads to unstable image synthesis, as the gradients derived from the emotional classifier may be adversarial to those generated by the text prompts. 
Inspired by~\cite{galimage}, we propose instead to optimize extra \textbf{emotional tokens} $S$ for emotion synthesis.

Specifically, as shown in Fig.~\ref{fig:main}, the original text prompt $P$ is embedded by a text encoder $\mathcal{T}$, and then concatenated with extra emotional tokens $S$,  to obtain the emotionalized text embedding $c^{emo}=\mathcal{T}(P) \oplus S$; here $ \oplus$ is concatenation of embeddings. Then, we compute the denoised embeddings $\hat{z}^{emo}_{t}$ conditioned on $c^{emo}$ by replacing $c$ in Eq.~\ref{eq: cfg}.

To well adapt an emotion classifier $\mathcal{D}_{emo}$ pre-trained with clean images to noisy data, we adopt the one-step inference similar to~\cite{bansal2023universal} to obtain the estimated clean embedding $\hat{z}^{emo}_{0}$ from each time step $t$ as follows:
\begin{equation}
\hat{z}^{emo}_{0} = \frac{\hat{z}^{emo}_t - \left(\sqrt{1 - \bar{\alpha}_{t}}\right) \epsilon_\theta\left(\hat{z}^{emo}_{t}, t, c^{emo}\right)}{\sqrt{\bar{\alpha}_{t}}},
\label{eq4:one_step_inference}
\end{equation}
where $\bar{\alpha}_{t}=\prod_{l=1}^t \alpha_l$. 
We then optimize emotional tokens $S$ via back-propagated gradients by minimizing the loss $\mathcal{L}_{\text{emo}}$ between the desired emotion $y_{\text{target}}$ and the predicted emotion $\hat{y}_{\text{emo}} {=} \mathcal{D}_{\text{emo}}(\hat{z}^{\text{emo}}_{0})$ during the denoising process, as:
\begin{equation}
S^{'}=\arg \min _{S} \mathcal{L}_{emo}\left(y_{\text {target}}, p\left(\hat{y}_{emo} \left\lvert\, \hat{z}^{emo}_{0}\right.\right)\right),
\label{eq:update_W}
\end{equation}
where $\mathcal{L}_{emo}$ is the target loss function detailed in Sec.~\ref{sec:which}. 

Our design of optimizing extra emotional tokens $S^{'}$ offers two main advantages: (1) It leverages prior knowledge from SOTA emotion classifiers directly, removing the need to create a large-scale emotional manipulation dataset. This is particularly beneficial given the subjectivity of emotions and the tremendous efforts required to create such a dataset. 
(2) Using both $S^{'}$ and $P$ together as framework inputs allows $\epsilon_{\theta}$ to simultaneously consider both semantic and emotional information from the prompts when estimating noise. This prevents gradient inconsistencies, avoids ignoring the textual information and ensures stable synthesis. 

Finally, since the emotional tokens $S$ only affect the model's input, there is no need for further complex operations on intermediate embeddings. The framework can be easily applied to both tasks by simply adapting the prompt and emotional token inputs  $(P, S)$ and noise samplings $z_{T}$.

\noindent
\textbf{Application in Generation:} 
Existing emotion generation methods like EmoGen~\cite{yang2024emogen} synthesize emotions only according to the target emotion, while our method can achieve it both with and without text descriptions:
For the generation with a text prompt, we set $P$ textual description (e.g. ``An elder woman cutting cake into slices") as in Fig.~\ref{fig:main} and randomly initialize tokens $S$. 
For the generation with only target emotion, we simply set $P=$ ``an image of ''+ target emotion. 
Both cases start from a random noise $z_{T} \sim \mathcal{N}(0, I)$.

\noindent
\textbf{Application in Editing:} 
For the editing task, we use the description of the source image to set $P$ and randomly initialize tokens $S$.
The noise $z_{T}$ is obtained from the source image by using the inversion in DDIM~\cite{songdenoising}.

\begin{figure*}[t]
    \centering
    \setlength{\tabcolsep}{1pt}
    \begin{tabular}{cc}
        \includegraphics[width=0.45\linewidth]{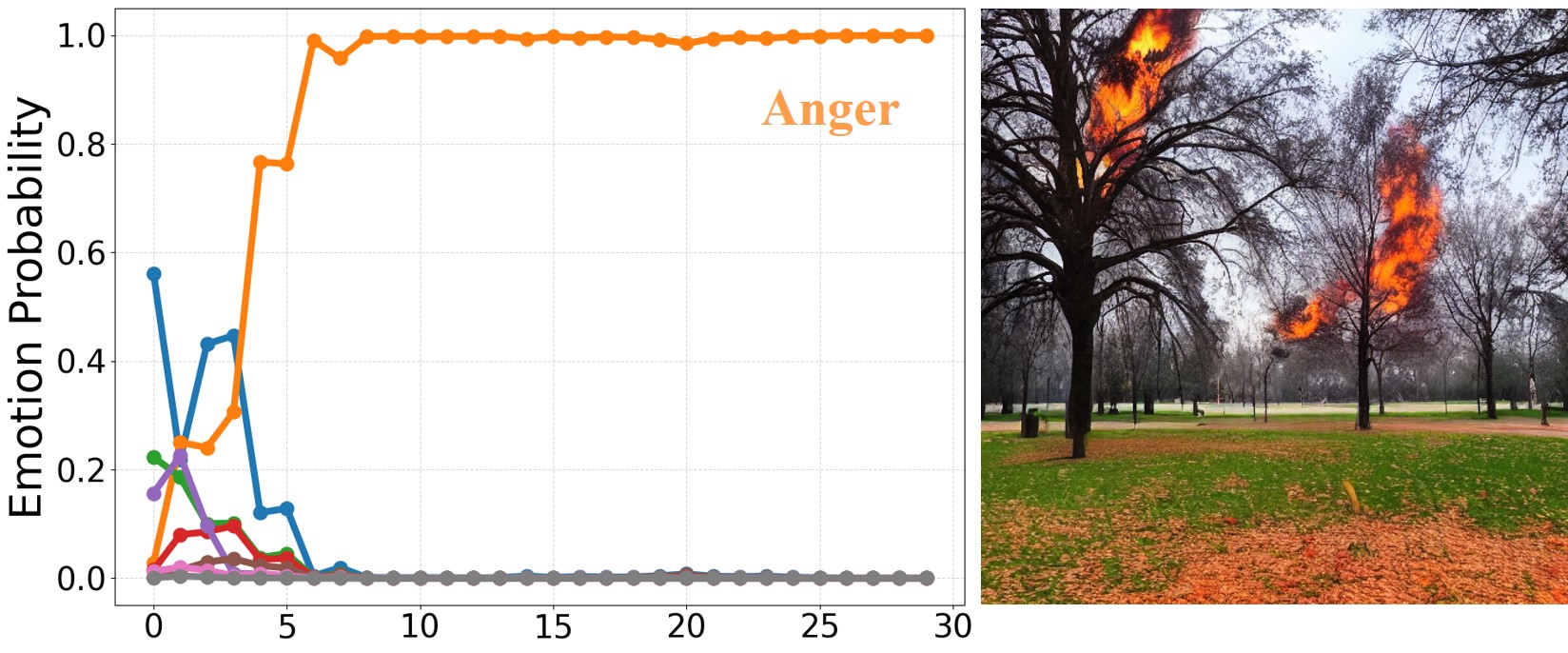} &
        \includegraphics[width=0.45\linewidth]{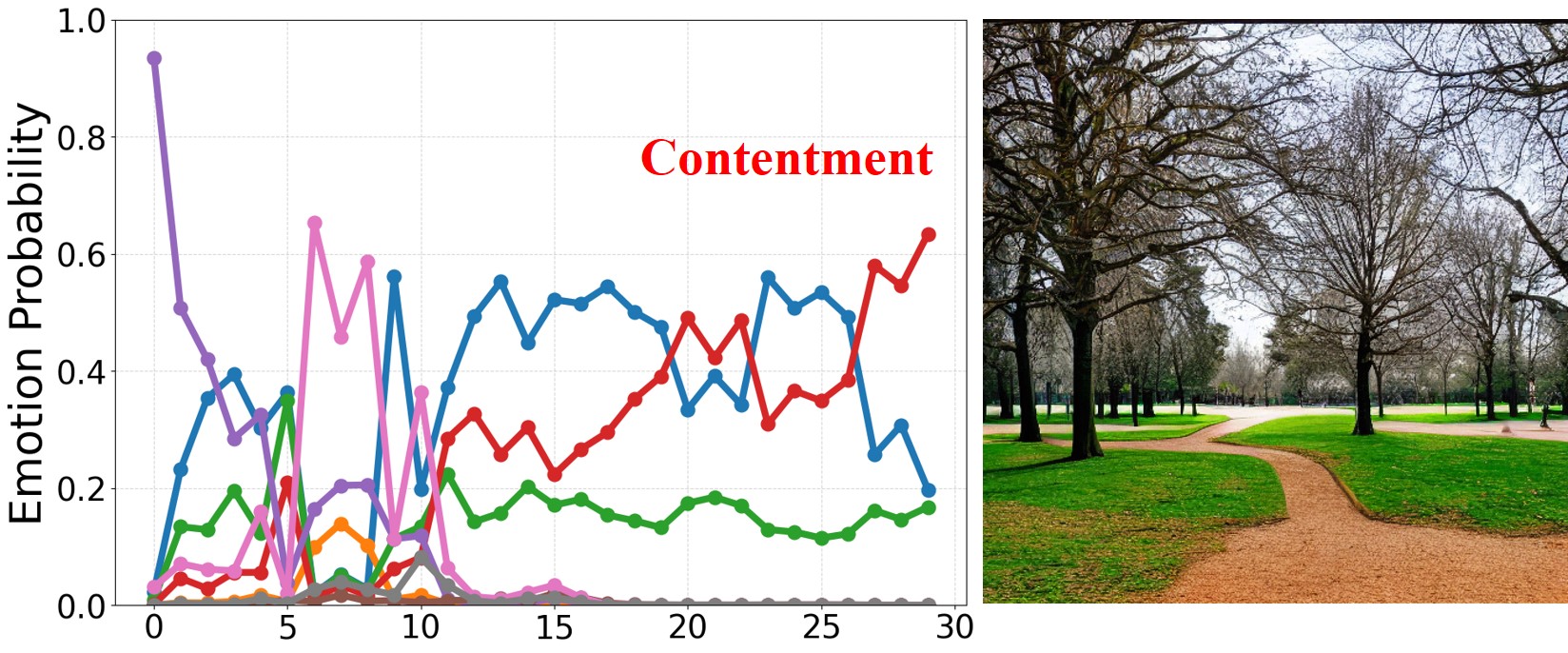} \\
        \footnotesize (a). Generated image controlled by inherent emotion``contentment" &
        \footnotesize (b). Generated image w.r.t. the target emotion ``anger" \\[2pt]

        \includegraphics[width=0.45\linewidth]{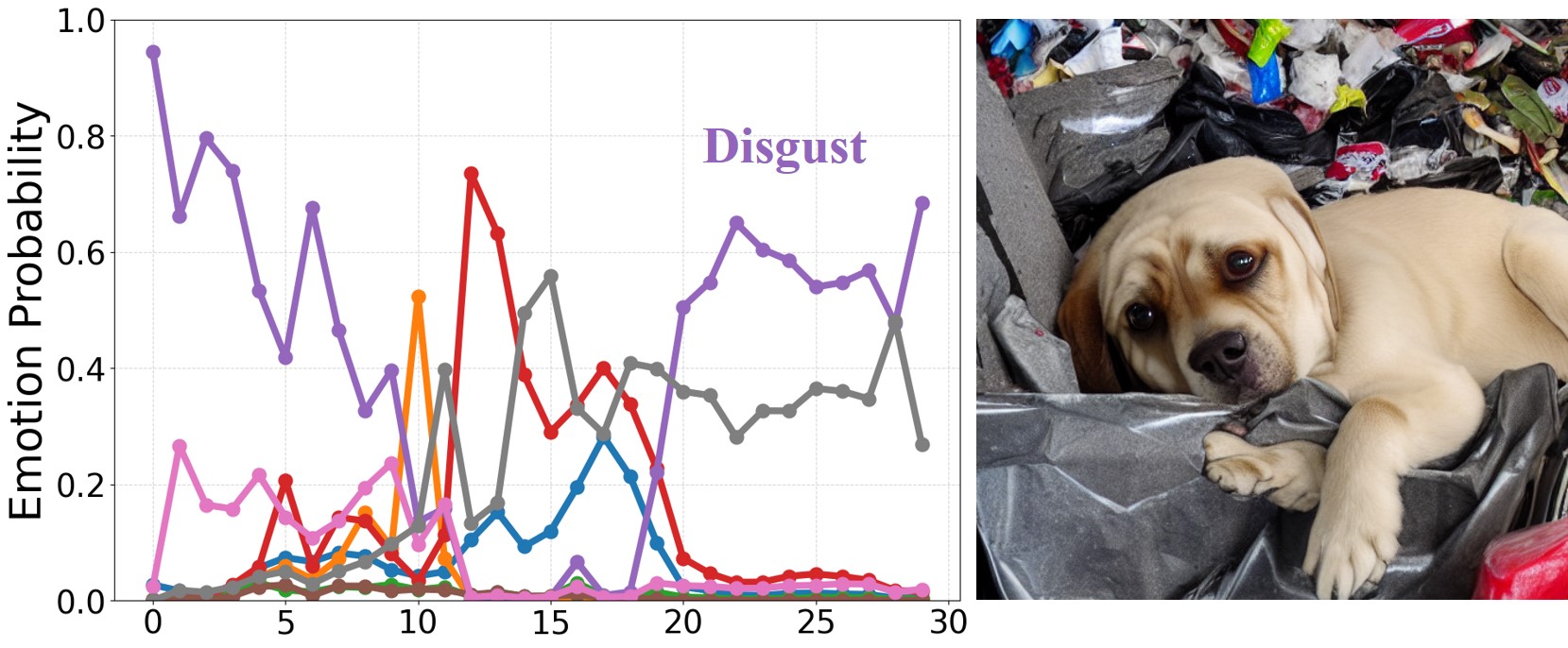} &
        \includegraphics[width=0.45\linewidth]{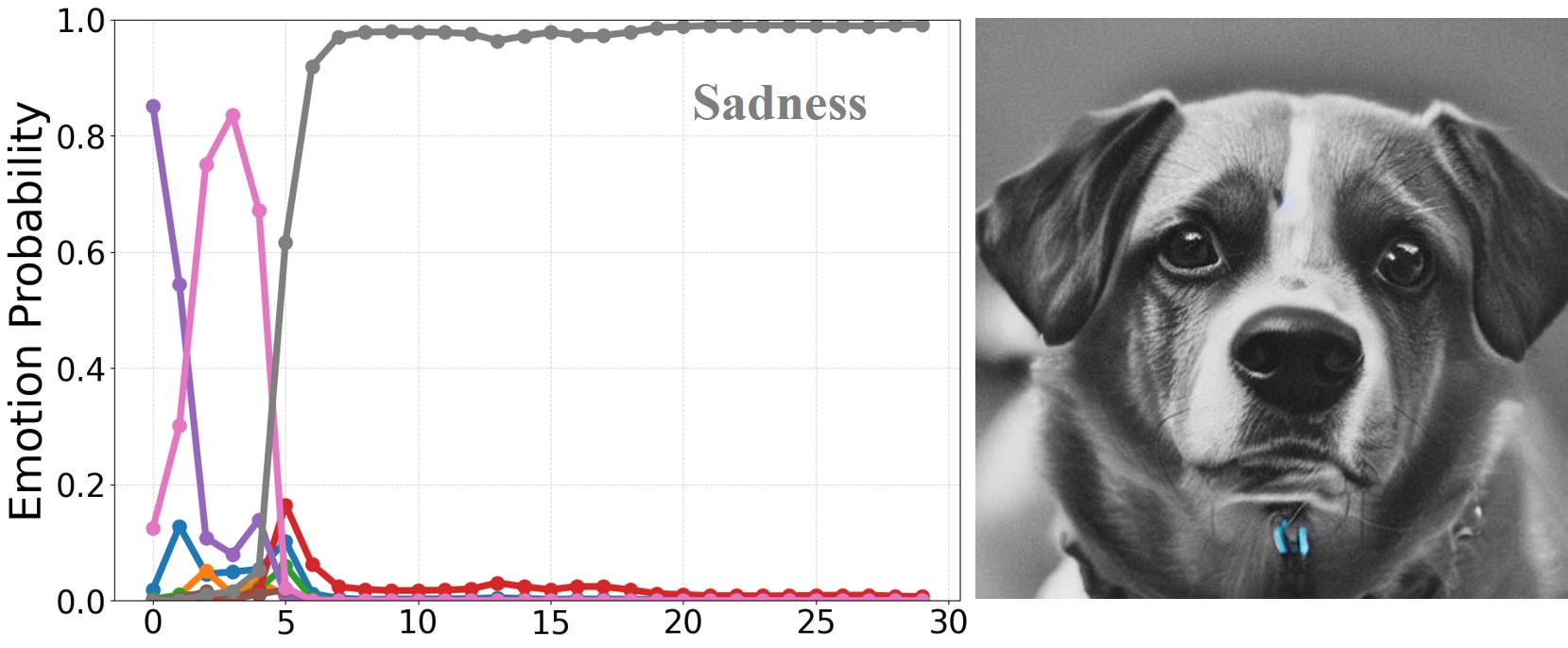} \\
        \footnotesize (c). Generated image misled by similar emotion ``disgust" &
        \footnotesize (d). Generated image w.r.t, the target emotion ``sadness" \\
    \end{tabular}

    \vspace{4pt}
    \includegraphics[width=0.88\linewidth]{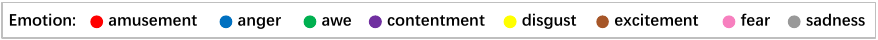}

    \caption{Visualization of suppressing similar and inherent emotions. (a) and (b) show the generation of a ``park" image w.r.t the target emotion ``anger", with and without inherent emotion suppression. (c) and (d) show the generation of a ``dog staring at you" image w.r.t the target emotion ``sadness", with and without similar emotion suppression. The x-axis denotes the number of inner emotional optimization loops, and the y-axis represents the emotion probability predicted by the classifier.}
    \label{fig:inherent_emotion}
\end{figure*}

\subsection{Supervision on Semantic Similarity}
\label{sec:when}
\textbf{\textit{When}} to start emotional guidance during the denoising is another challenge for emotion synthesis: 
\begin{itemize}
    \item  \textbf{Adding guidance too early.} During the early stage of denoising, the image content is {not yet determined~\cite{hertz2023prompt}}. Using both the text prompt $P$ and the optimized emotional tokens $S^{'}$ as inputs may cause the network to focus excessively on the target emotion but neglect the semantics. 
    The final image has an accurate target emotion but risks semantic deviation.

    \item \textbf{Adding guidance too late.} In the later stages of denoising, where the main content of the image is almost decided, the model tends to concentrate only on details such as texture and subtle structures. 
    At this point, strong inherent emotions present in the textual prompts or image content (e.g. an image of a park during the day has the inherent emotion of ``contentment") may risk incorporating also the target emotion. 
    This results in the final image with accurate semantic but emotional deviation.
\end{itemize}

Introducing the guidance of the target emotion at the \textbf{right timing} can suppress inherent emotions while preserving semantic accuracy.
To achieve this without increasing the complexity of the denoising, we leverage the image encoder $\mathcal{E}^{\text{clip}}_{I}$ and text encoder $\mathcal{E}^{\text{clip}}_{T}$ from CLIP~\cite{radford2021learning}
to measure the semantic similarity $s^{\text{clip}}_{t}$ (with a temperature factor $\tau=0.07$) between the generated content and text prompts at each time step $t$: 
\begin{equation}
s^{clip}_{t} = \frac{1}{\tau}\text{CLIP}(\mathcal{E}^{\text{clip}}_{I}(\mathcal{D}(\hat{z}_{t\to 0})),\mathcal{E}^{\text{clip}}_{T}(P)),
\end{equation}

where $\hat{z}_{t\to 0}$ is the embedding iteratively derived from $\hat{z}_{t}$ to the time step $t=0$ by using Eq.~\ref{eq3:denoise_process}.
In the reverse process, once $s^{\text{clip}}_{t}$ exceeds a preset threshold $\eta$, it indicates that the image content has preliminarily formed, then the emotional guidance $S$ is optimized and introduced:
\begin{equation}
c^{\text{emo}}=\begin{cases} \mathcal{T}(P) \oplus \varnothing & s^{clip}_{t}<\eta \\ \mathcal{T}(P) \oplus S^{'} & s^{clip}_{t}>=\eta
\end{cases},
\label{eq:threshold}
\end{equation}
with $\varnothing$ is a void placeholder. This mechanism enables us to introduce target emotional guidance at the right stage, ensuring the convey of the target emotion.

\begin{figure*}[!ht]
 \centering
  \includegraphics[width=0.81\textwidth]{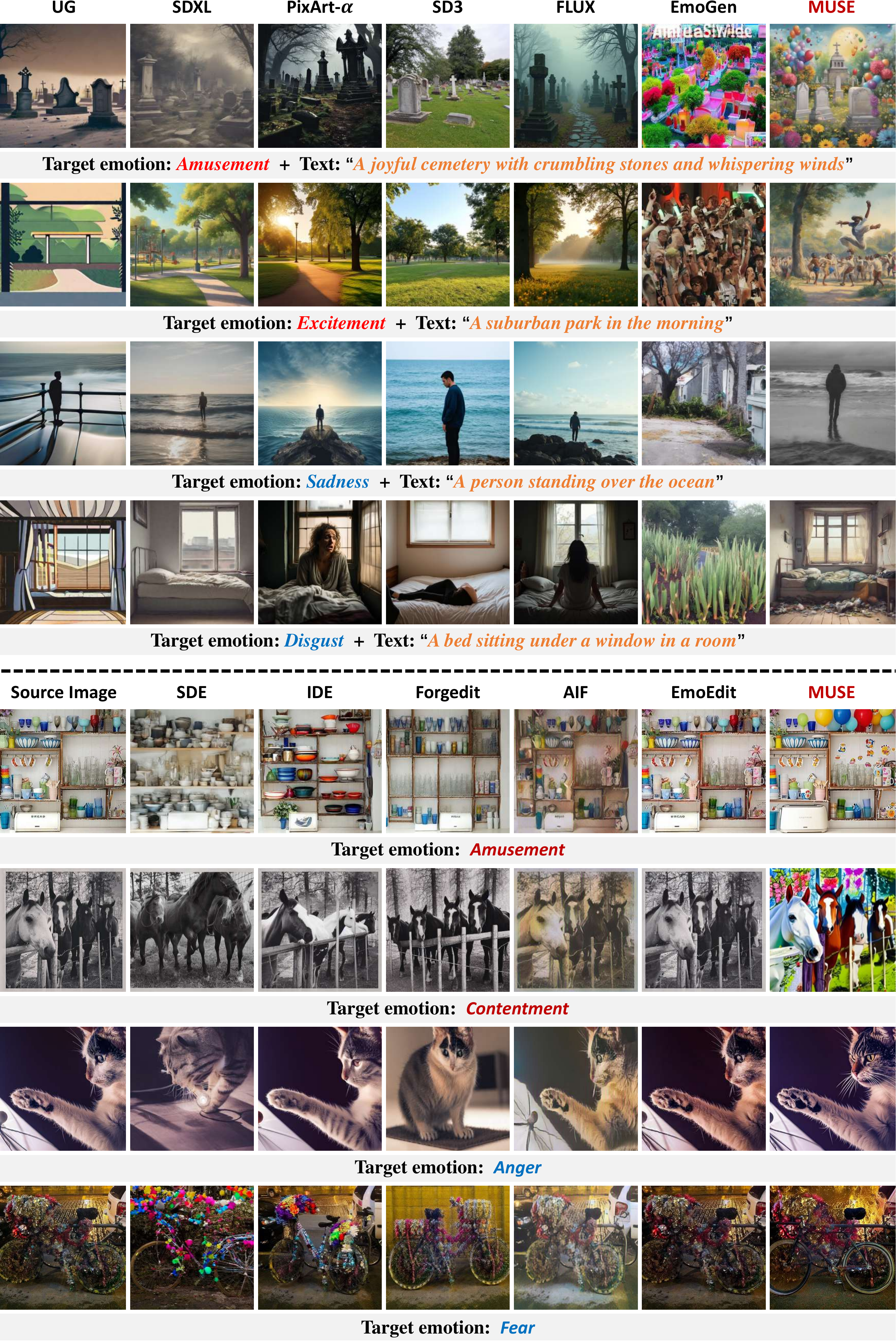}
  \caption{
  Qualitative comparison of emotional image generation and editing methods. 
\textbf{Upper Part}: Emotion generation results from SD~\cite{rombach2022high}, UG~\cite{bansal2023universal}, PixArt-$\alpha$~\cite{chen2024pixart}, SDXL~\cite{rombach2022high}, SD3~\cite{esser2024scaling}, FLUX~\cite{flux2024}, and EmoGen~\cite{yang2024emogen}. UG and MUSE are conditioned on both text prompts and emotion labels, while other methods rely solely on emotionally descriptive prompts. 
\textbf{Lower Part}: Emotion editing results from SDE~\cite{mengsdedit}, AIF~\cite{weng2023affective}, IDE~\cite{zou2024towards}, Forgedit~\cite{zhang2023forgedit}, and EmoEdit~\cite{yang2025emoedit}. All editing methods take a neutral input image and modify it according to the target emotion.}
  \label{fig:gallery}
\end{figure*}

\begin{table*}[ht]\footnotesize
\centering
\scalebox{0.9}
{
\begin{tabular}{ccccccccccccccccccccc}
\toprule[0.4pt]
Backbone &\multicolumn{2}{l}{{Methods}} & \multicolumn{2}{c}{Emo-A ($\%$)$\uparrow$} & \multicolumn{2}{c}{Emo-B ($\%$) $\uparrow$} & \multicolumn{2}{c|}{Emo-C ($\%$)$\uparrow$}  & \multicolumn{6}{c|}{FID $\downarrow$}  & \multicolumn{2}{c}{CLIP $\uparrow$} &\multicolumn{2}{c}{HPSV2 $\uparrow$}\\

&\multicolumn{2}{c}{} & \multicolumn{2}{c}{} & \multicolumn{2}{c}{} & \multicolumn{2}{c|}{} &\multicolumn{2}{c}{\cellcolor{blue!25}{EmoSet~\cite{yang2023emoset}}}    & \multicolumn{2}{c}{\cellcolor{red!25}{FI\_8}~\cite{you2016building}} & \multicolumn{2}{c|}{\cellcolor{green!25}{COCO~\cite{lin2014microsoft}}} \\ 

\midrule[0.4pt]
Transformer-Unet &\multicolumn{2}{l}{PixArt-$\alpha$~\cite{chen2024pixart}} &\multicolumn{2}{c}{23.87}    & \multicolumn{2}{c}{25.29} & \multicolumn{2}{c|}{25.54} & \multicolumn{2}{c}{63.16} & \multicolumn{2}{c}{63.90} & \multicolumn{2}{c|}{51.91} &\multicolumn{2}{c}{30.33} & \multicolumn{2}{c}{0.22}\\
\midrule[0.4pt]
SD3.0 &\multicolumn{2}{l}{SD3~\cite{esser2024scaling}} &\multicolumn{2}{c}{18.28}    & \multicolumn{2}{c}{21.07} & \multicolumn{2}{c|}{21.41} & \multicolumn{2}{c}{76.68} & \multicolumn{2}{c}{79.70}  & \multicolumn{2}{c|}{42.21} &\multicolumn{2}{c}{31.51}  & \multicolumn{2}{c}{0.18}\\
\midrule[0.4pt]
FLUX1.0 &\multicolumn{2}{l}{FLUX~\cite{flux2024}} &\multicolumn{2}{c}{18.93}    & \multicolumn{2}{c}{21.80} & \multicolumn{2}{c|}{20.73} & \multicolumn{2}{c}{80.67}  & \multicolumn{2}{c}{82.53} &\multicolumn{2}{c|}{52.23} & \multicolumn{2}{c}{30.76}  & \multicolumn{2}{c}{0.24} \\
\midrule[0.4pt]
\multirow{5}{*}{SD1.5} & \multicolumn{2}{l}{SD~\cite{rombach2022high}} &\multicolumn{2}{c}{15.00}    & \multicolumn{2}{c}{15.81} & \multicolumn{2}{c|}{16.55} & \multicolumn{2}{c}{64.34} & \multicolumn{2}{c}{67.18}  & \multicolumn{2}{c|}{\underline{41.02}} &\multicolumn{2}{c}{\underline{28.72}}  & \multicolumn{2}{c}{0.18} \\

&\multicolumn{2}{l}{UG~\cite{bansal2023universal}} &\multicolumn{2}{c}{23.46}    & \multicolumn{2}{c}{\underline{33.10}} & \multicolumn{2}{c|}{18.78} & \multicolumn{2}{c}{67.04} & \multicolumn{2}{c}{62.87} & \multicolumn{2}{c|}{61.68} &\multicolumn{2}{c}{23.73}  & \multicolumn{2}{c}{\underline{0.21}}\\

&\multicolumn{2}{l}{EmoGen~\cite{yang2024emogen}} &\multicolumn{2}{c}{\underline{30.96}}    & \multicolumn{2}{c}{28.72} & \multicolumn{2}{c|}{\underline{28.69}} & \multicolumn{2}{c}{\underline{43.80}} & \multicolumn{2}{c}{\underline{45.96}}  & \multicolumn{2}{c|}{59.11} &\multicolumn{2}{c}{23.39}  & \multicolumn{2}{c}{0.17}\\

&\multicolumn{2}{l}{MUSE} &\multicolumn{2}{c}{\textbf{\maxf{68.38}}}    & \multicolumn{2}{c}{\textbf{\maxf{59.38}}} & \multicolumn{2}{c|}{\textbf{\maxf{32.23}}} & \multicolumn{2}{c}{\textbf{\maxf{43.53}}} & \multicolumn{2}{c}{\textbf{\maxf{44.18}}}  & \multicolumn{2}{c|}{\textbf{\maxf{26.52}}} &\multicolumn{2}{c}{\textbf{\maxf{30.33}}}  & \multicolumn{2}{c}{\textbf{\maxf{0.24}}}\\

\midrule[0.4pt]

\multirow{3}{*}{SDXL1.0} &\multicolumn{2}{l}{SDXL~\cite{podellsdxl}}&\multicolumn{2}{c}{\underline{19.48}}    & \multicolumn{2}{c}{\underline{21.50}} & \multicolumn{2}{c|}{\underline{20.35}} & \multicolumn{2}{c}{\underline{69.54}} & \multicolumn{2}{c}{\textbf{\maxf{69.42}}}  & \multicolumn{2}{c|}{\underline{39.76}} &\multicolumn{2}{c}{\underline{30.59}}  & \multicolumn{2}{c}{\textbf{\maxf{0.26}}}\\

&\multicolumn{2}{l}{UG~\cite{bansal2023universal}} &\multicolumn{2}{c}{18.03}    & \multicolumn{2}{c}{17.45} & \multicolumn{2}{c|}{16.10} & \multicolumn{2}{c}{131.66} & \multicolumn{2}{c}{120.74} & \multicolumn{2}{c|}{126.14} &\multicolumn{2}{c}{26.14}  & \multicolumn{2}{c}{0.18}\\

&\multicolumn{2}{l}{MUSE} &\multicolumn{2}{c}{\textbf{\maxf{41.87}}}   & \multicolumn{2}{c}{\textbf{\maxf{39.87}}} & \multicolumn{2}{c|}{\textbf{\maxf{30.14}}} & \multicolumn{2}{c}{\textbf{\maxf{69.30}}} & \multicolumn{2}{c}{\underline{69.87}}  & \multicolumn{2}{c|}{\textbf{\maxf{38.28}}} &\multicolumn{2}{c}{\textbf{\maxf{31.23}}}  & \multicolumn{2}{c}{\underline{0.24}}\\

\bottomrule[0.4pt]
\end{tabular}}
\caption{\textbf{Comparison against the SOTA T2I emotion generation methods.} For all methods, when accessible, we evaluate both the SD~\cite{rombach2022high} and SDXL~\cite{podellsdxl} backbones. 
Both UG, EmoGen and MUSE take emotion as a label for supervision, while SD, SDXL, SD3 and FLUX textualize emotion as the prompt as in PixArt-$\alpha$~\cite{chen2024pixart}.
We use 3 classifiers to measure accuracy: Emo-A with the classifier pretrained by EmoSet~\cite{yang2023emoset}, Emo-B with the training-agnostic classifier from ~\cite{yang2024emogen}, and Emo-C with the same classifier as Emo-A, but pretrained on FI\_8~\cite{you2016building}.
We also measure FID on 3 datasets, COCO~\cite{lin2014microsoft}, Emoset~\cite{yang2023emoset}, and FI\_8~\cite{you2016building}.
We further resepcitvely quantify the text-image adherence and human preference with CLIP score (CLIP) and HPSV2~\cite{wu2023human}. 
The best performance is highlighted in \textbf{bold} with a gray background, and the second-best performance is \underline{underlined}. 
}
\label{tab:quantitative_generation}
\end{table*}

 \begin{table}[htbp]
\centering
\resizebox{\linewidth}{!}{
\begin{tabular}{cccccccccccc}
\toprule
\multicolumn{2}{l}{{Methods}} &\multicolumn{2}{c}{Emo-A ($\%$)$\uparrow$} &\multicolumn{2}{c}{Emo-B ($\%$)$\uparrow$} &\multicolumn{2}{c|}{Emo-C ($\%$)$\uparrow$} & \multicolumn{2}{c}{CLIP $\uparrow$} &\multicolumn{2}{c}{HPSV2 $\uparrow$} \\ 
\midrule[0.8pt]
\multicolumn{2}{l}{SDE~\cite{mengsdedit}} & \multicolumn{2}{c}{14.80}   & \multicolumn{2}{c}{15.24} & \multicolumn{2}{c|}{15.49} & \multicolumn{2}{c}{0.78} & \multicolumn{2}{c}{\underline{0.23}}\\
\multicolumn{2}{l}{PnP~\cite{tumanyan2023plug}} & \multicolumn{2}{c}{13.37}     & \multicolumn{2}{c}{13.50} & \multicolumn{2}{c|}{13.63} & \multicolumn{2}{c}{\textbf{\maxf{0.86}}} & \multicolumn{2}{c}{\underline{0.23}}\\

\multicolumn{2}{l}{AIF~\cite{weng2023affective}} & \multicolumn{2}{c}{12.44} & \multicolumn{2}{c}{12.44} & \multicolumn{2}{c|}{12.56} & \multicolumn{2}{c}{\textbf{\maxf{0.86}}} & \multicolumn{2}{c}{0.21}\\

\multicolumn{2}{l}{Forgedit~\cite{zhang2023forgedit}} & \multicolumn{2}{c}{13.50} & \multicolumn{2}{c}{13.19} & \multicolumn{2}{c|}{13.69} & \multicolumn{2}{c}{0.76} & \multicolumn{2}{c}{\underline{0.23}}\\

\multicolumn{2}{l}{IDE~\cite{zou2024towards}} & \multicolumn{2}{c}{23.87} & \multicolumn{2}{c}{25.29} & \multicolumn{2}{c|}{14.04} & \multicolumn{2}{c}{\underline{0.83}} & \multicolumn{2}{c}{\textbf{\maxf{0.24}}}\\

\multicolumn{2}{l}{EmoEdit~\cite{yang2025emoedit}} &\multicolumn{2}{c}{\underline{29.31}}    & \multicolumn{2}{c}{\underline{29.13}} & \multicolumn{2}{c|}{\underline{27.44}} & \multicolumn{2}{c}{\textbf{\maxf{0.86}}} & \multicolumn{2}{c}{0.22}\\
\midrule[0.4pt]
\multicolumn{2}{l}{MUSE} & \multicolumn{2}{c}{\textbf{\maxf{69.50}}} & \multicolumn{2}{c}{\textbf{\maxf{59.86}}} & \multicolumn{2}{c|}{\textbf{\maxf{34.43}}} & \multicolumn{2}{c}{0.74} & \multicolumn{2}{c}{\underline{0.23}}\\
\bottomrule
\end{tabular}
}
\caption{Comparisons with the SOTA methods on the editing task. In all cases, the input condition is an image and the emotion label.}
\label{tab:abs_sa}
\end{table}

\subsection{Suppression of Undesired Emotions}
\label{sec:which}
\textbf{\textit{Which}} emotion to guide the synthesis should be emphasized throughout the generation. Two types of undesired emotions affect the expression of the target emotion: 

\noindent \textbf{(1) Inherent Emotions} derived from the source image or text prompt. E.g., an image or text related to ``park" naturally carries the inherent emotion ``contentment". 
Once semantic content is primarily formed, this inherent emotion continuously tries to take control of the expressed emotion (see Fig.~\ref{fig:inherent_emotion}(a)). 
To suppress the inherent emotion, we propose applying the semantic-similarity supervision mechanism from Sec.~\ref{sec:when}, obtaining the embedding $\hat{z}_{t'}$ at the time step $t'$ before $s^{\text{clip}}_{t}$ exceeds the threshold $\eta$ (i.e. $t'{=}t{+}1$). 
We then predict the inherent emotion $y_{\text{inh}}$ from $\hat{z}_{t'}$ with the emotional classifier, and apply a cross-entropy loss to suppress the emotional conflict:
\begin{equation}
\mathcal{L}_{\text{inh}} =\ell_{\text{ce}}(y_{\text{inh}},p(\hat{y}_{\text{emo}} \lvert\, \hat{z}^{\text{emo}}_{0})),
\end{equation}
where $\ell_{\text{ce}}$ is a cross-entroy loss. This loss term can suppress the inherent emotion in synthesized images (see Fig.~\ref{fig:inherent_emotion}(b)).

\noindent \textbf{(2) Similar Emotions} usually relate to the target emotion. Unlike other classification tasks where categories have distinct and mutually exclusive characteristics, categories of emotion exhibit relative similarity~\cite{zhao2024to}. 
For example, ``disgust" and ``sadness" are similar emotions positioned adjacent to each other in Mikel's wheel psychological model~\cite{mikels2005emotional}. 
Similar emotions can lead to a shift in the guided emotion toward them during the synthesis (see Fig.~\ref{fig:inherent_emotion}(c)). 
The similar emotions $y_\text{{sim}}$ associated with the target emotion $y_{\text{target}}$ can be obtained by querying the emotion wheel~\cite{mikels2005emotional}, and they can be suppressed using cross-entropy loss:
\begin{equation}
\mathcal{L}_{\text{sim}} =\ell_{\text{ce}}(y_{\text{sim}},p(\hat{y}_{\text{emo}} \lvert\, \hat{z}^{\text{emo}}_{0})).
\end{equation}

In summary, to suppress these two types of undesired emotion and to force moving towards the target emotion, we propose the following multi-emotion loss used in Eq.~\ref{eq:update_W}:
\begin{equation}
\mathcal{L}_{\text{\text{emo}}}= \mathcal{L}_{\text{target}} -\lambda_{1}\mathcal{L}_{\text{inh}}-\lambda_{2}\mathcal{L}_{\text{sim}},
\label{eq:loss_function}
\end{equation}
where $\mathcal{L}_{\text{target}}=\max(0,\ell_{\text{ce}}(y_{\text{target}},p(\hat{y}_{\text{emo}} \lvert\, \hat{z}^{emo}_{0}))$ is loss in terms of target emotion, and $\lambda_{1}$, $\lambda_{2}$ are hyperparameters to control the suppression strength. 
During each step, an inner-loop optimization is performed on the emotional token, and ended once the loss $\mathcal{L}_{\text{\text{emo}}}$ is lower than a preset threshold $10^{-4}$.

\section{Experiments}
\label{sec:exp}

\subsection{Dataset and Evaluation}
\label{dataset_evaluation}
\noindent
\textbf{Datasets:} We carry out experiments on three datasets: EmoSet~\cite{yang2023emoset}, COCO~\cite{lin2014microsoft}, and FI\_8~\cite{you2016building}.
EmoSet~\cite{yang2023emoset} consists of 118,102 images of eight emotion classes. The off-the-shelf classifier we use is pre-trained on this dataset.
Following~\cite{rombach2022high,podellsdxl}, we also employ COCO~\cite{lin2014microsoft} due to its diversity and detailed text descriptions by sampling image-text pairs from it. 
FI\_8~\cite{you2016building} is used as an external validation set to evaluate image fidelity, which contains 22,697 images. 

\noindent
\textbf{Metrics \& Settings:}
For the generation task, we assess image fidelity using FID~\cite{heusel2017gans} and emotion accuracy with Emo-A, Emo-B and Emo-C. 
Emo-A is measured using EmoSet-pretrained SimEmo~\cite{deng2022simple} classifier, while Emo-B involves a classifier directly from~\cite{yang2024emogen} to fairly show generalization in emotional accuracy. 
And Emo-C also measured with the same classifier as used by Emo-A, but pre-trained on FI\_8~\cite{you2016building}.
We use the CLIP score~\cite{radford2021learning} to evaluate text-image alignment. 
And we use HPSV2~\cite{wu2023human} to assess image quality and aesthetics. 
For the editing task, we use Emo-A, Emo-B, Emo-C, CLIP, and HPSV2. 
We use the pre-trained Stable Diffusion v1.5\footnote{\url{https://huggingface.co/runwayml/stable-diffusion-v1-5}}, Stable Diffusion XL v1.0\footnote{\url{https://huggingface.co/stabilityai/stable-diffusion-xl-base-1.0}} and CLIP model\footnote{\url{https://huggingface.co/openai/clip-vit-large-patch14}} as model backbone and semantic similarity measurement.
We adopt the Adam optimizer~\cite{kingma2014adam} with a linearly decaying learning rate \( lr = 0.01 \cdot \left(1 - \frac{t}{100} \right) \). 
To balance the loss terms, we set \( \lambda_1 = 0.0005 \) and \( \lambda_2 = 0.0015 \) for \( \mathcal{L}_{\text{inh}} \) and \( \mathcal{L}_{\text{sim}} \).

\subsection{Comparison to the state of the art}
For the generation task, we compare MUSE against backbones Stable Diffusion (SD)~\cite{rombach2022high}, SDXL~\cite{podellsdxl}, SD3~\cite{esser2024scaling} and FLUX~\cite{flux2024}, as well as one T2I synthesis SOTA method PixArt-\(\alpha\)~\cite{chen2024pixart}. 
These methods explicitly incorporate the target emotion through textual prompts.
We also benchmark against Universal Guidance (UG)~\cite{bansal2023universal}, and EmoGen~\cite{yang2024emogen}, a SOTA emotional image generation method.
For editing task, 
we compare MUSE with representative methods such as SDEit (SDE)~\cite{mengsdedit}, PnP~\cite{tumanyan2023plug}, Forgedit~\cite{zhang2023forgedit}, InstDiffEdit (IDE)~\cite{zou2024towards},  AIF~\cite{weng2023affective} and EmoEdit~\cite{yang2025emoedit}.
And we built MUSE on SD and SDXL backbones for both tasks.
We adopted the categorization of emotion from~\cite{mikels2005emotional}, which include four positive (amusement, awe, contentment, excitement) and four negative emotions (anger, disgust, fear, and sadness).
We randomly sampled text captions from the COCO dataset~\cite{lin2014microsoft}, which provides diverse textual descriptions. 

\noindent \textbf{Emotional Generation: }
For the emotional generation task, it's crucial to align both the textual semantic, image content and target emotion.
We generated 700 images per emotion (totally 5600 images) with MUSE and evaluate its performance with six metrics.
We demonstrate quantitative and qualitative comparisons in Table~\ref{tab:quantitative_generation} and upper part of Fig.~\ref{fig:gallery}, respectively.
We can observe from Tab.~\ref{tab:quantitative_generation} that MUSE built on both SD and SDXL backbones achieves the highest emotional evocation accuracy (Emo-A, B \& C), surpassing the 2nd-best methods by about 30\%, 20\% and 10\%.
Moreover, it also proves the robustness of MUSE as it achieves the best accuracy with a training classifier (Emo-A), an agnostic classifier (Emo-B), and a cross-validated classifier (Emo-C). 
MUSE also gives the best FID for SD across all datasets, while MUSE-SDXL ranks 1st in two out of three datasets, with FI\_8 slightly lagging behind. 
Universal Guidance~\cite{bansal2023universal} shows the worst FID, and EmoGen~\cite{yang2024emogen} exhibits degraded FID on COCO, likely due to overfitting to its emotional dataset. 
Both SD and SDXL deliver strong FIDs, consistent with their high-quality backbones.
In terms of adherence, MUSE outperforms others on the CLIP metric, while methods like UG~\cite{bansal2023universal} and EmoGen~\cite{yang2024emogen} struggle with the text-emotion alignment, often overly focusing either on the text description or the desired emotion in the synthesized images (see Fig.~\ref{fig:gallery}). 
Additionally, as measured by HPSv2~\cite{wu2023human}, MUSE also exhibits good performance in human aesthetic evaluations.

The upper section of Fig.~\ref{fig:gallery} demonstrates MUSE's superior capability in aligning emotional expression with semantic content. For example, in the 2nd row, our method successfully evokes the target emotion of ``excitement" by incorporating ``dancing crowds," while baseline methods merely focus on depicting the literal ``park" scene. More notably, when handling semantic-emotional conflicts (1st row) where the concept of ``cemetery" inherently contradicts ``amusement" or ``joyful" emotions, MUSE effectively introduces vibrant visual elements to convey the desired emotional tone. In contrast, comparison methods exhibit significant limitations: PixArt-$\alpha$~\cite{chen2024pixart} and SDXL~\cite{podellsdxl} completely fail to express the target emotion, while EmoGen~\cite{yang2024emogen} deviates from the original prompt semantics.

\begin{figure*}[!ht]
    \centering
    \setlength{\tabcolsep}{1pt}
    \begin{tabular}{cccc}
        \includegraphics[width=0.24\linewidth]{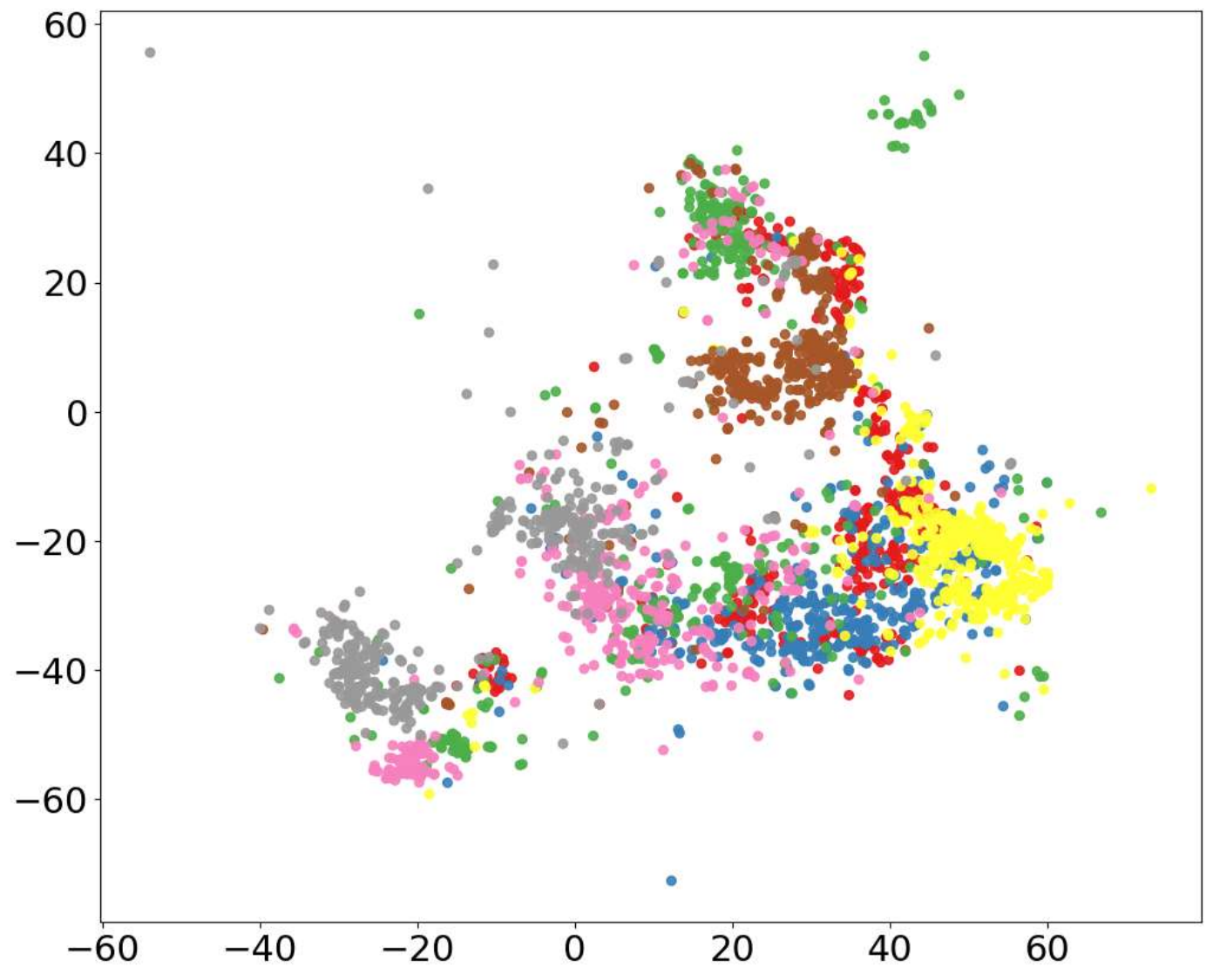} &
        \includegraphics[width=0.24\linewidth]{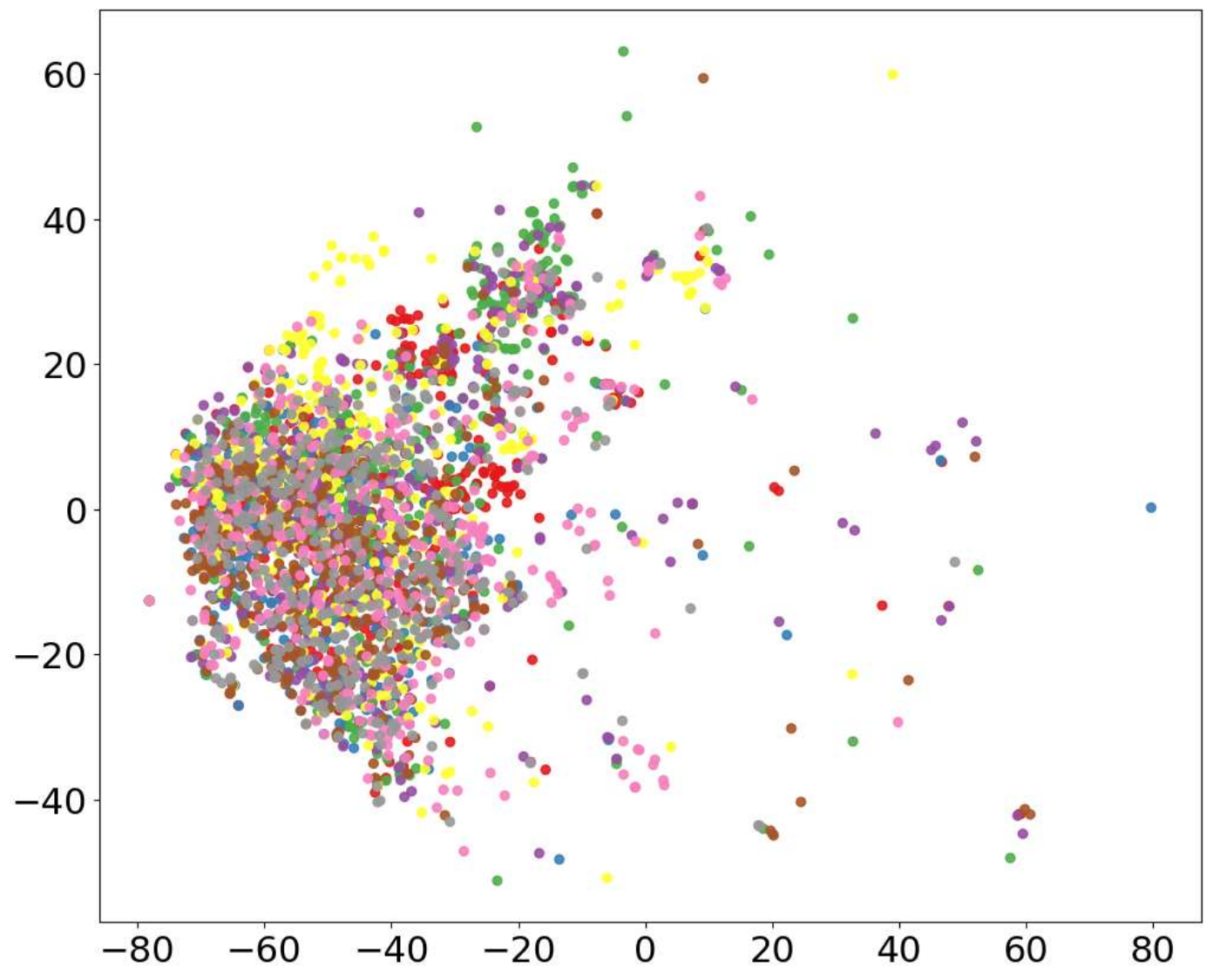} &
        \includegraphics[width=0.24\linewidth]{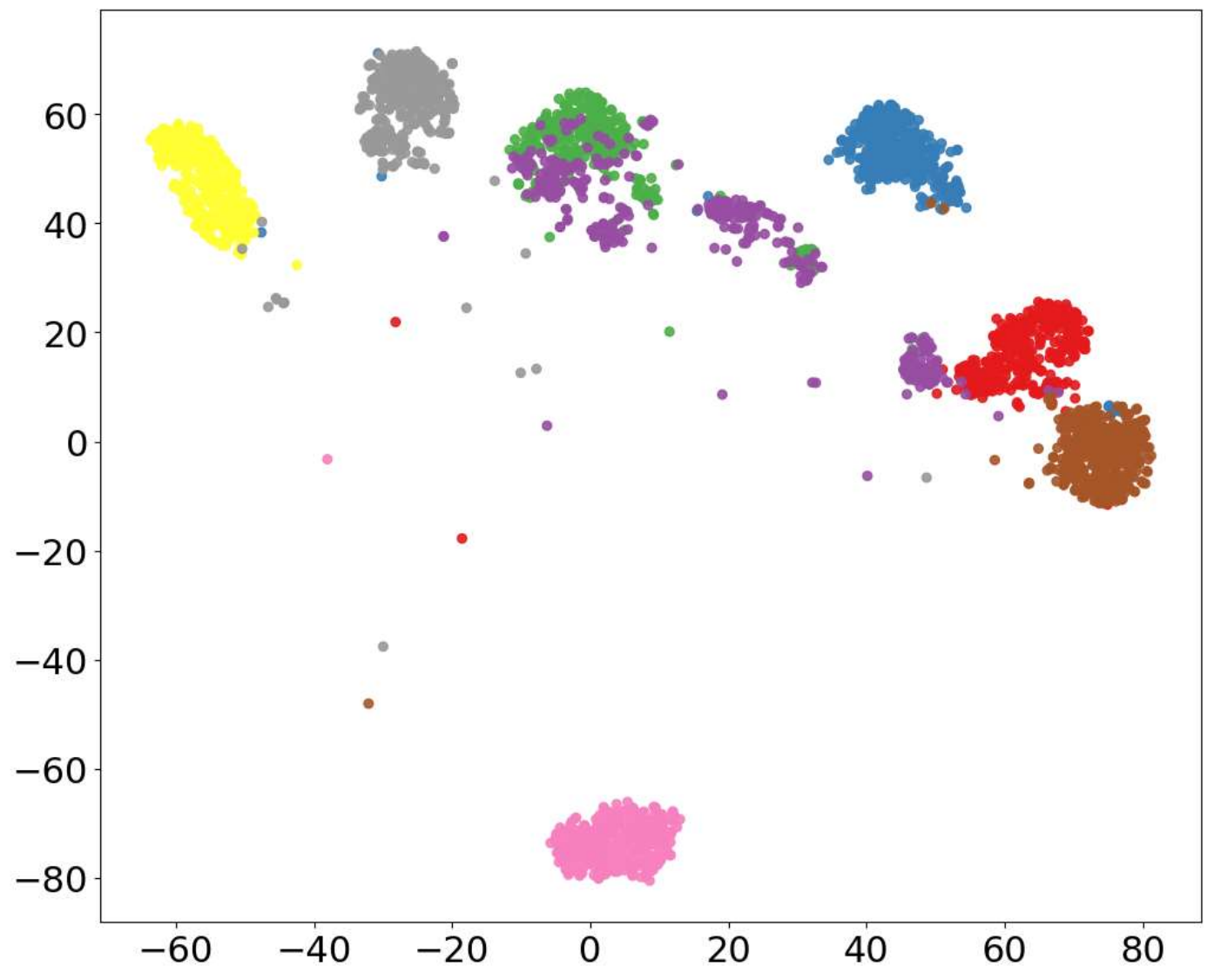} &
        \includegraphics[width=0.24\linewidth]{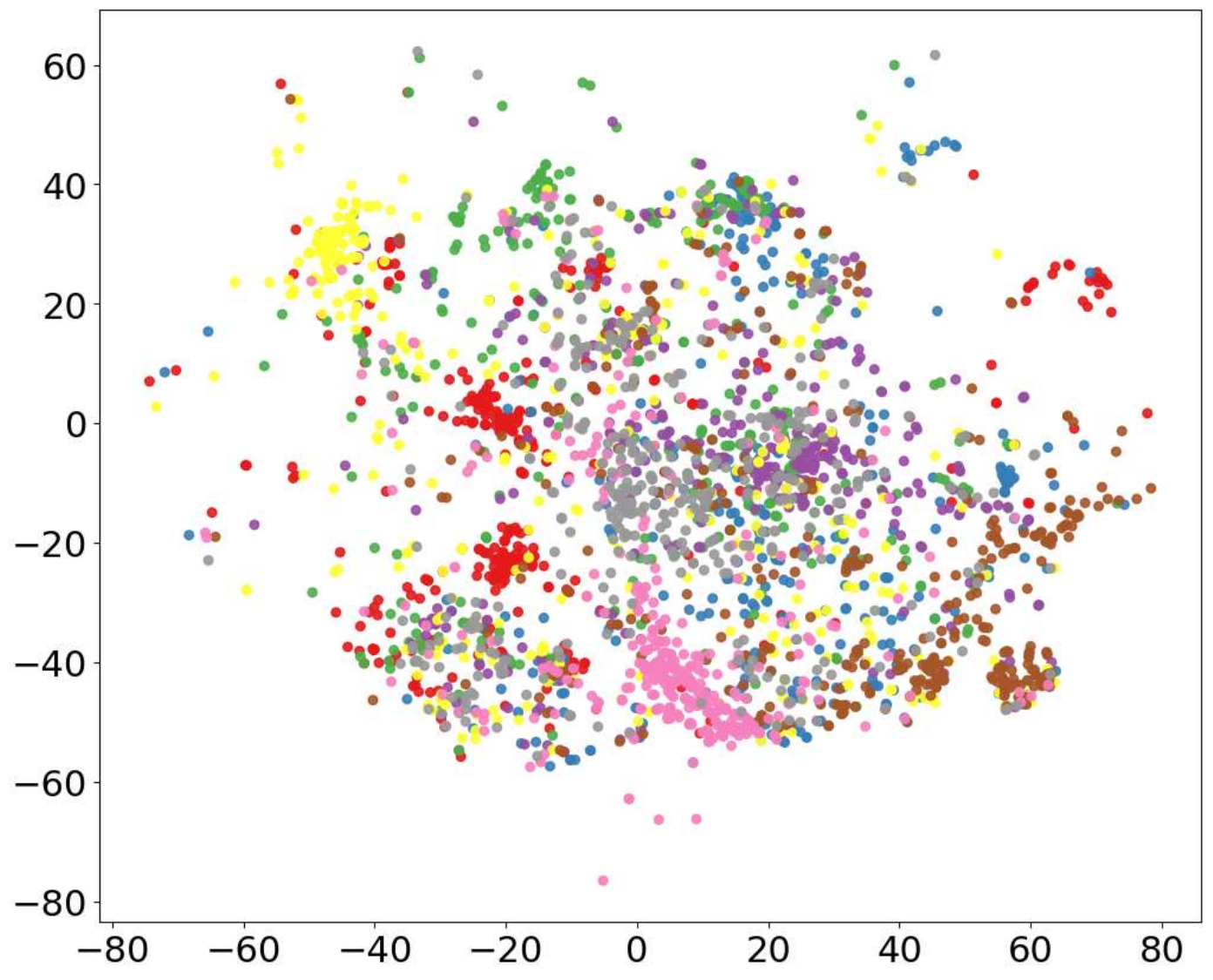} \\
         \footnotesize \textit{Emo-A($\%$): 38.84, FID: 60.69 }&
        \footnotesize \textit{Emo-A($\%$): 50.98, FID: 85.17} &
        \footnotesize \textit{Emo-A($\%$): 74.55, FID: 73.61} &
        \footnotesize \textit{Emo-A($\%$): 83.71, FID: 49.23}\\       
        \footnotesize   \textit{Var: 0.101,} SD~\cite{rombach2022high}&
        \footnotesize   \textit{Var: 0.072,} UG~\cite{bansal2023universal} &
        \footnotesize   \textit{Var: 0.069,} EmoGen~\cite{yang2024emogen}&
        \footnotesize   \textit{Var: 0.103,} MUSE\\
    \end{tabular}

    \vspace{4pt}
    \includegraphics[width=0.88\linewidth]{figures/sup/emotion_label.pdf}

    \caption{T-SNE~\cite{van2008visualizing} visualization of feature distributions for non-textual emotional generation across four models. Each data point represents the CLIP semantic features of an image, with different colors indicating their corresponding emotion categories. The Var is the averaged intra-class variance of CLIP features. 
    Compared to the other three methods (a), (b), and (c), MUSE achieves better separation of content within the same emotion category while exhibiting a more dispersed overall semantic distribution.}
    \label{fig:tsne_map}
\end{figure*}

\begin{figure*}[t]
\centering
\begin{minipage}{0.68\textwidth}
\resizebox{\linewidth}{!}{
\begin{tabular}{ccc cccccc cccccc cc}
\toprule[0.4pt]
\multicolumn{1}{l}{Guidance}  & Loss:  & Optim & \multicolumn{2}{c}{Emo-A} & \multicolumn{2}{c}{Emo-B} & \multicolumn{2}{c|}{Emo-C}   & \multicolumn{6}{c|}{FID $\downarrow$}  & \multicolumn{2}{c}{CLIP}\\

\multicolumn{1}{l}{Classifier} & $\mathcal{L}_{\text{target}}$ - & ET$^{\dag}$  & \multicolumn{2}{c}{($\%$)$\uparrow$} & \multicolumn{2}{c}{($\%$)$\uparrow$} & \multicolumn{2}{c|}{($\%$)$\uparrow$}  &\multicolumn{2}{c}{\cellcolor{blue!25}{EmoSet}}    & \multicolumn{2}{c}{\cellcolor{red!25}{FI\_8}} & \multicolumn{2}{c|}{\cellcolor{green!25}{COCO}} & \multicolumn{2}{c}{$\uparrow$} \\ 

\midrule[0.4pt]

\multicolumn{1}{l}{EmoGen~\cite{yang2024emogen}} & $\mathcal{L}_\text{sim}$ - $\mathcal{L}_\text{inh}$  &  \checkmark &\multicolumn{2}{c}{55.38}   & \multicolumn{2}{c}{\textbf{\maxf{72.38}}} & \multicolumn{2}{c|}{\textbf{\maxf{33.42}}} & \multicolumn{2}{c}{\underline{44.55}} & \multicolumn{2}{c}{46.20}  & \multicolumn{2}{c|}{26.92} &\multicolumn{2}{c}{29.86} \\

\midrule[0.4pt]

\multicolumn{1}{l}{SimEmo$^{*}$~\cite{deng2022simple}}    & / &  \checkmark &\multicolumn{2}{c}{66.78}    & \multicolumn{2}{c}{58.05} & \multicolumn{2}{c|}{30.91} & \multicolumn{2}{c}{44.66} & \multicolumn{2}{c}{\underline{44.82}} & \multicolumn{2}{c|}{\underline{26.64}} &\multicolumn{2}{c}{29.34}\\

\multicolumn{1}{l}{SimEmo$^{*}$~\cite{deng2022simple}} &  $\mathcal{L}_\text{sim}$ &  \checkmark  &\multicolumn{2}{c}{\underline{67.05}}    & \multicolumn{2}{c}{57.01} & \multicolumn{2}{c|}{31.57} & \multicolumn{2}{c}{45.04} & \multicolumn{2}{c}{45.53} & \multicolumn{2}{c|}{29.40} &\multicolumn{2}{c}{\textbf{\maxf{30.42}}} \\

\multicolumn{1}{l}{SimEmo$^{*}$~\cite{deng2022simple}} & $\mathcal{L}_\text{inh}$  & \checkmark &\multicolumn{2}{c}{67.04}    & \multicolumn{2}{c}{58.23} & \multicolumn{2}{c|}{31.95} & \multicolumn{2}{c}{45.76} & \multicolumn{2}{c}{46.39} & \multicolumn{2}{c|}{26.83} &\multicolumn{2}{c}{30.20}  \\

\multicolumn{1}{l}{SimEmo$^{*}$~\cite{deng2022simple}}   & $\mathcal{L}_\text{sim}$ - $\mathcal{L}_\text{inh}$  & \xmark &\multicolumn{2}{c}{12.94}   & \multicolumn{2}{c}{12.72} & \multicolumn{2}{c|}{13.26} & \multicolumn{2}{c}{60.55} & \multicolumn{2}{c}{67.43}  & \multicolumn{2}{c|}{42.37} &\multicolumn{2}{c}{30.00} \\

\midrule

\multicolumn{1}{l}{SimEmo$^{*}$~\cite{deng2022simple}}  & $\mathcal{L}_\text{sim}$ - $\mathcal{L}_\text{inh}$  & \checkmark &\multicolumn{2}{c}{\textbf{\maxf{68.38}}}    & \multicolumn{2}{c}{\underline{59.38}} & \multicolumn{2}{c|}{\underline{32.23}} & \multicolumn{2}{c}{\textbf{\maxf{43.53}}} & \multicolumn{2}{c}{\textbf{\maxf{44.18}}}  & \multicolumn{2}{c|}{\textbf{\maxf{26.52}}} &\multicolumn{2}{c}{\underline{30.33}}  \\

\bottomrule[0.4pt]
\multicolumn{8}{c}{$^{\dag}$ OET indicates Optimizing (or not) the Emotional Token. $*$ means classifier trained by ourselves.}
\end{tabular}
}
\captionof{table}{
\textbf{Ablation study}. The 1st row shows the performance of MUSE using the classifier from EmoGen~\cite{yang2024emogen}, the 2nd-4th rows show the results of MUSE trained using different loss term combinations, and the 5th row is the results of MUSE when removing the emotional token optimization. The final row shows results using our proposed MUSE setting.
} 
\label{tab:ablation_study}
\end{minipage}
\hfill%
\begin{minipage}{0.3\textwidth}
  \centering
  
  \includegraphics[width=0.95\linewidth]{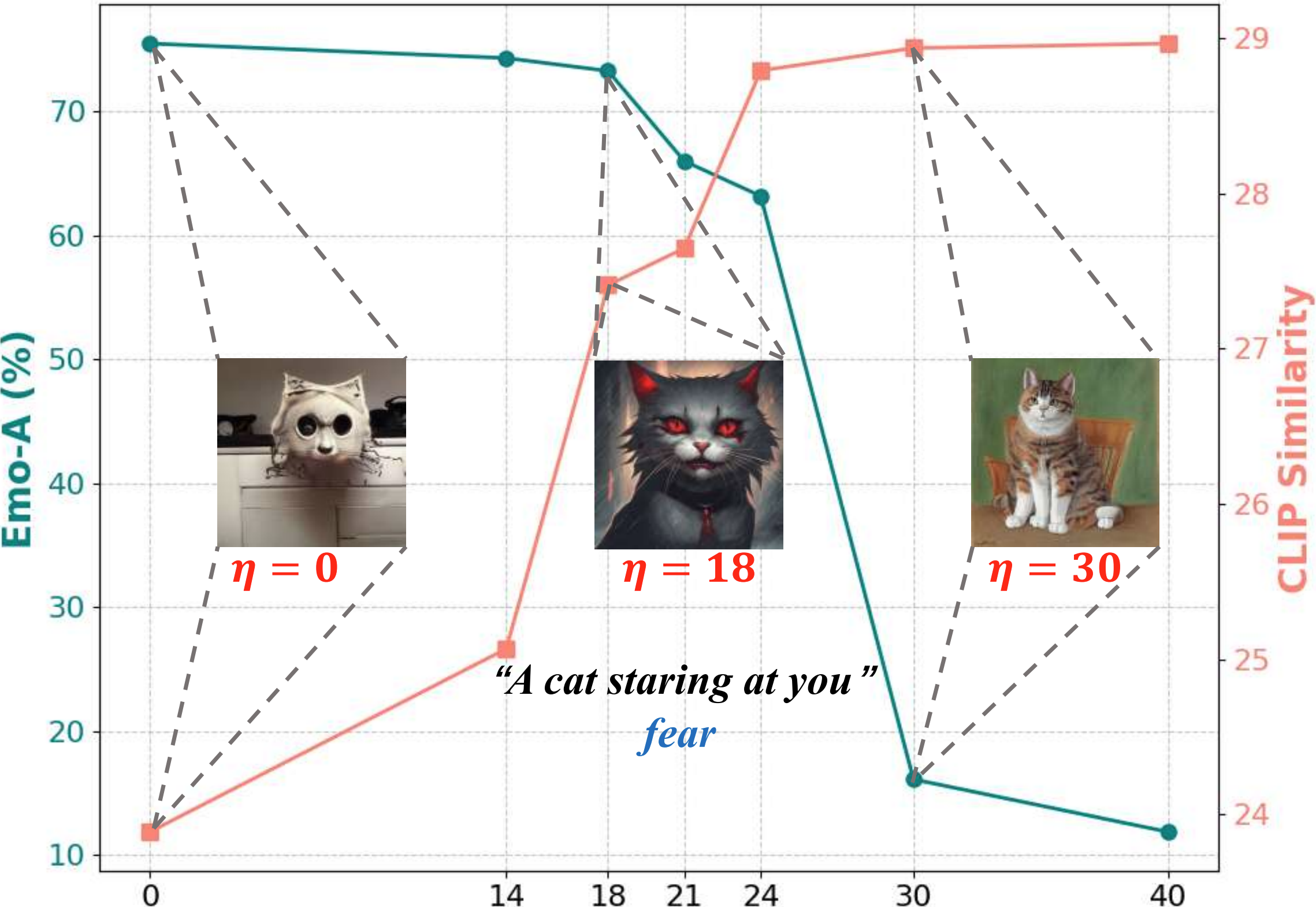}
     \caption{Evolution of accuracy \& similarity in terms of threshold $\eta$. Images shown are generated at early, optimal, and late stages of emotional guidance. 
     }
     \label{fig:timing-ablation}
\end{minipage}
\end{figure*}

\noindent
\textbf{Emotional Editing:}
For the emotional editing task, we randomly select 200 image-text pairs per emotion (totally 1.6K images) from COCO~\cite{lin2014microsoft}. 
We compare MUSE with SOTA approaches, including global editing methods SDEdit~\cite{mengsdedit} and PnP~\cite{tumanyan2023plug}, the style-based method AIF~\cite{weng2023affective}, and the emotion-specific editing method EmoEdit~\cite{yang2025emoedit}. 
Additionally, we include editing methods based on the diffusion model such as Forgedit~\cite{zhang2023forgedit} and InstDiffEdit (IDE)~\cite{zou2024towards}.
The results are shown in Table~\ref{tab:abs_sa} and the lower part of Fig.~\ref{fig:gallery}. 
We observe that AIF~\cite{weng2023affective} PnP~\cite{tumanyan2023plug} and EmoEdit~\cite{yang2025emoedit} achieve high CLIP similarity of 0.86, but suffer from emotion accuracy lower than 14\%. 
EmoEdit~\cite{yang2025emoedit} performs better, yet its emotional accuracy remains below 30\%. 
IDE~\cite{zou2024towards} aligns better images with target emotions with emotional accuracy around 24\%, but often introduces irrelevant elements (e.g., adding a vase on a cabinet or ribbons on a bicycle in the last row of Fig.~\ref{fig:gallery}). 
In contrast, MUSE produces more natural edits by balancing image content and emotions, achieving notably higher emotion accuracies (at least 7\% accuracy gain) and a competitive aesthetic score of 0.23, despite the slightly lower CLIP score, thus highlighting its editing capabilities.

\begin{table*}[t]\footnotesize
\centering
\scalebox{1.0}
{
    \begin{tabular}{cccccccccccccccccccc}
    \toprule[0.4pt]
    \multicolumn{2}{l}{{Guidance}} & \multicolumn{2}{c}{Emo-A ($\%$)$\uparrow$} & \multicolumn{2}{c}{Emo-B ($\%$) $\uparrow$} & \multicolumn{2}{c|}{Emo-C ($\%$)$\uparrow$}  & \multicolumn{6}{c|}{FID $\downarrow$}  & \multicolumn{2}{c}{CLIP $\uparrow$}  &\multicolumn{2}{c}{HPSV2 $\uparrow$}\\

    \multicolumn{2}{l}{Timing} & \multicolumn{2}{c}{} & \multicolumn{2}{c}{} & \multicolumn{2}{c|}{}  &\multicolumn{2}{c}{\cellcolor{blue!25}{EmoSet}}    & \multicolumn{2}{c}{\cellcolor{red!25}{FI\_8}} & \multicolumn{2}{c|}{\cellcolor{green!25}{COCO}}  \\
    \midrule[0.05pt]

    \multicolumn{2}{l}{${\eta=0}$} &\multicolumn{2}{c}{\textbf{\maxf{74.30}}}   & \multicolumn{2}{c}{\textbf{\maxf{60.85}}} & \multicolumn{2}{c|}{\textbf{\maxf{36.62}}} & \multicolumn{2}{c}{88.64} & \multicolumn{2}{c}{84.91}  & \multicolumn{2}{c|}{88.92} &\multicolumn{2}{c}{23.11}  & \multicolumn{2}{c}{\underline{0.23}} \\
    
    \multicolumn{2}{l}{${\eta=14}$} &\multicolumn{2}{c}{\underline{73.26}}    & \multicolumn{2}{c}{\underline{59.60}} & \multicolumn{2}{c|}{\underline{35.76}} & \multicolumn{2}{c}{69.46} & \multicolumn{2}{c}{\underline{66.50}} & \multicolumn{2}{c|}{44.53} &\multicolumn{2}{c}{23.90}  & \multicolumn{2}{c}{0.22} \\

    \multicolumn{2}{l}{${\eta=18}$} &\multicolumn{2}{c}{68.38}    & \multicolumn{2}{c}{59.38}& \multicolumn{2}{c|}{32.23} & \multicolumn{2}{c}{\textbf{\maxf{43.53}}} & \multicolumn{2}{c}{\textbf{\maxf{44.18}}}  & \multicolumn{2}{c|}{\textbf{\maxf{26.52}}} &\multicolumn{2}{c}{\underline{30.33}}  & \multicolumn{2}{c}{\textbf{\maxf{0.24}}} \\

    \multicolumn{2}{l}{${\eta=21}$} &\multicolumn{2}{c}{65.97}   & \multicolumn{2}{c}{55.22} & \multicolumn{2}{c|}{31.63}& \multicolumn{2}{c}{\underline{62.11}} & \multicolumn{2}{c}{68.55}  & \multicolumn{2}{c|}{51.10} &\multicolumn{2}{c}{30.08}  & \multicolumn{2}{c}{\underline{0.23}} \\
    
    \multicolumn{2}{l}{${\eta=24}$} &\multicolumn{2}{c}{63.16}   & \multicolumn{2}{c}{51.22} & \multicolumn{2}{c|}{27.38}& \multicolumn{2}{c}{68.40} & \multicolumn{2}{c}{67.78}  & \multicolumn{2}{c|}{50.25} &\multicolumn{2}{c}{\textbf{\maxf{30.41}}}  & \multicolumn{2}{c}{\underline{0.23}} \\
    
    \multicolumn{2}{l}{${\eta=30}$} &\multicolumn{2}{c}{16.09}   & \multicolumn{2}{c}{14.02} & \multicolumn{2}{c|}{13.80}& \multicolumn{2}{c}{68.19} & \multicolumn{2}{c}{67.43}  & \multicolumn{2}{c|}{\underline{42.37}} &\multicolumn{2}{c}{30.00}  & \multicolumn{2}{c}{\underline{0.23}} \\
    \bottomrule[0.4pt]
    \end{tabular}}
\vspace{-0.2cm}
\caption{Generation ablation on $\eta$ compared with SOTA methods.}
\label{tab:threshold}
\end{table*}

\noindent
\textbf{Semantic Visualization}
To provide a more direct observation of the semantics for generated images,  
we employed T-SNE~\cite{van2008visualizing} to visualize in Fig.~\ref{fig:tsne_map} the distribution of 3200 generated images (400 images per emotion) in the CLIP semantic space for methods SD~\cite{rombach2022high}, UG~\cite{bansal2023universal} and EmoGen~\cite{yang2024emogen} and measure their corresponding averaged intra-class variance.
EmoGen~\cite{yang2024emogen} explicitly maps each emotion to specific attributes, resulting in a semantic space that clearly displays several emotional clusters. This structure implies that images sharing the same emotion tend to exhibit similar visual elements, leading to reduced semantic diversity.
On the other hand, SD~\cite{rombach2022high} and UG~\cite{bansal2023universal} reveal different characteristics. SD~\cite{rombach2022high} prioritizes the preservation of semantic diversity and image quality but tends to overlook emotion. This approach leads to higher variance and FID scores, while yielding lower emotional accuracy.
In contrast, UG~\cite{bansal2023universal} places greater emphasis on emotion, at the expense of semantic diversity and image quality.
Our MUSE model combines the strengths of SD~\cite{rombach2022high}, achieving high FID and diversity, while also attaining superior emotional accuracy without explicitly mapping emotions to attributes, as in EmoGen~\cite{yang2024emogen}.

\begin{table}[!h]
\centering
\resizebox{\linewidth}{!}{ 
\begin{tabular}{cccccccccccc}
\toprule
\multicolumn{2}{l}{{Methods}} &\multicolumn{2}{c}{Qua $\downarrow$} &\multicolumn{2}{c}{Sem $\downarrow$} & \multicolumn{2}{c}{Emo $\downarrow$} &\multicolumn{2}{c}{Div $\downarrow$} &\multicolumn{2}{c}{Adh $\downarrow$} \\ 

\midrule[0.8pt]
\multicolumn{2}{l}{SD~\cite{rombach2022high}} & \multicolumn{2}{c}{\textbf{\maxf{2.45}}}   & \multicolumn{2}{c}{\underline{2.45}}  & \multicolumn{2}{c}{2.83} & \multicolumn{2}{c}{\underline{2.50}} & \multicolumn{2}{c}{2.89} \\
\multicolumn{2}{l}{UG~\cite{bansal2023universal}} & \multicolumn{2}{c}{2.53}     & \multicolumn{2}{c}{2.62} & \multicolumn{2}{c}{2.71} & \multicolumn{2}{c}{2.60}  & \multicolumn{2}{c}{2.54} \\
\multicolumn{2}{l}{EmoGen~\cite{yang2024emogen}} & \multicolumn{2}{c}{2.52} & \multicolumn{2}{c}{2.62} & \multicolumn{2}{c}{\underline{2.56}} & \multicolumn{2}{c}{2.83} & \multicolumn{2}{c}{\underline{2.31}} \\
\midrule[0.4pt]
\multicolumn{2}{l}{MUSE} & \multicolumn{2}{c}{\underline{2.50}} & \multicolumn{2}{c}{\textbf{\maxf{2.31}}} & \multicolumn{2}{c}{\textbf{\maxf{1.90}}} & \multicolumn{2}{c}{\textbf{\maxf{2.07}}} & \multicolumn{2}{c}{\textbf{\maxf{2.26}}} \\
\bottomrule
\end{tabular}}
\caption{User study, numbers are the average rank.}
\label{tab:user_study}
\end{table}

\noindent \textbf{User Study:}
We generated 160 images (8 emotions {$\times$} 5 text prompts) for SD~\cite{rombach2022high}, UG~\cite{bansal2023universal}, EmoGen~\cite{yang2024emogen} and MUSE. 
In a blind evaluation, 15 participants ranked the images across five dimensions: 
image quality (Qua), semantic fidelity (Sem), emotional fidelity (Emo), content diversity (Div), and adherence between emotion and semantics (Adh). 
Results are shown in Tab.~\ref{tab:user_study}. 
Although MUSE slightly trails SD~\cite{rombach2022high} in image quality, it achieves the highest average rank in other criteria, especially emotional fidelity and diversity. 
This demonstrates that MUSE provides a more engaging visual experience and stronger emotional impact for viewers.

\subsection{Ablation Study}
\noindent
\textbf{Emotion Guidance Timing:} 
To prove the importance of emotion guidance timing, i.e. when to start emotional token optimization, we randomly sample 200 textual prompts from the COCO dataset, generating one image per prompt for 8 emotions. 
We test different thresholds $\eta{=}\{0, 14, 18, 21, 24, 30\}$ for generation and show results in Tab.~\ref{tab:threshold}.
Among them, $\eta{=}\{0,14\}$ and $\eta{=}\{24,30\}$ means early and late stages of guidance introduction respectively. 
Fig.~\ref{fig:timing-ablation} shows the evolution of emotion accuracy and CLIP scores with respect to $\eta$. 
Early guidance produces correct emotions but incorrect semantics, while late guidance shows inverse results. 
Only the guidance at the right timing (e.g. $\eta{=}\{18, 21\}$) yields a good balance between emotions and semantics.
And $\eta{=}18$ gives the best balance of emotion accuracy and image quality.
We also remove emotional token optimization (5th row in Tab.~\ref{tab:ablation_study}), which is equivalent to $\eta{=}50$, where MUSE fails to synthesize correct emotions entirely.

\noindent
\textbf{Emotional Loss Term: }
To validate the effectiveness of our emotional loss in suppressing inherent and similar emotions, we test different combinations of loss terms in Eq.~\ref{eq:loss_function} (see 2nd-4th rows in Tab.~\ref{tab:ablation_study}). 
The results show that adding $\mathcal{L}_{\text{inh}}$ and $\mathcal{L}_{\text{sim}}$ reduces interference from inherent and similar emotions, improving the final emotion evocation accuracy.

\noindent
\textbf{Guidance Classifier:} 
To further investigate whether our method can effectively leverage the emotional priors learned by the classifier and to explore the influence of dataset and architecture, we conducted experiments on different classifiers, as shown in Tab.~\ref{tab:ablation_study}.
Specifically, we refer to the pretrained EmoGen~\cite{yang2024emogen} classifier as \textit{EmoGen}, and to an identical classifier architecture adopted from~\cite{deng2022simple} but pretrained on the same EmoSet dataset as \textit{SimEmo}.
For metrics, Emo-A and Emo-B are evaluated with the \textit{EmoGen}~\cite{yang2024emogen} and \textit{SimEmo}~\cite{deng2022simple} classifiers, whereas Emo-C is assessed using a \textit{SimEmo} model pretrained on the smaller FI\_8 dataset (see Sec.~\ref{dataset_evaluation}).

A comparison between the first (\textit{EmoGen}) and last (\textit{SimEmo}) rows of Tab.~\ref{tab:ablation_study} shows that either classifier can effectively steer emotional‐content synthesis. 
Nonetheless, the accuracy metrics (Emo-A/B) are consistently favored toward whichever classifier supplies the guidance signal, exposing certain architecture-dependent bias. 
These results illustrate the generality of our framework on different classifiers while also motivating evaluation under multiple architectures; accordingly, we adopt \textit{SimEmo} as the default guide in our experiments.
Finally, when the classifiers for guidance and testing share the same architecture but are pretrained on different datasets, MUSE demonstrates higher Emo-C than other methods (as shown in Tab.~\ref{tab:quantitative_generation}), though the knowledge gap (Emo-C is trained under a smaller dataset) between the datasets leads to a significant drop in performance.

Given this inherent knowledge gap in datasets, methods like MUSE that can quickly leverage priors from each dataset without the need for retraining exhibit a clear advantage in emotional synthesis.

\section{Conclusion}
We presented a unified framework for image emotion synthesis and editing, addressing three key aspects: how, when, and which emotion to be manipulated during synthesis. 
Specifically, we leverage the prior knowledge of external emotion classifiers and introduce a test-time optimization strategy during inference. 
Reverse optimization of emotional tokens is applied to maintain synthesis stability, while semantic similarity is used to determine the precise timing of emotional injection, ensuring semantic alignment. 
In addition, an multi-emotion loss function is employed to suppress undesired emotional interference and alleviate emotional conflicts. 
Experimental results demonstrate superior performance in both emotional accuracy and visual quality compared to existing methods for both task. 
Future work will focus on reducing knowledge gap between dataset and inference time. 
Moreover, as real-world emotional expression encompasses a far richer range than the eight discrete classes, leaving subtler or compound affects will also be a meaningful direction of our work.

\bibliographystyle{IEEEtran}
\bibliography{IEEEabrv,main}
\end{document}